\documentclass{article} % For LaTeX2e
\usepackage{iclr2027_conference,times}

\usepackage{amsmath,amsfonts,bm}

\def\eqref#1{equation~\ref{#1}}
\def\1{\bm{1}}

\def\mA{{\bm{A}}}

\def\mH{{\bm{H}}}
\def\mI{{\bm{I}}}

\def\mK{{\bm{K}}}
\def\mL{{\bm{L}}}

\def\mQ{{\bm{Q}}}

\def\mX{{\bm{X}}}

\DeclareMathAlphabet{\mathsfit}{\encodingdefault}{\sfdefault}{m}{sl}
\SetMathAlphabet{\mathsfit}{bold}{\encodingdefault}{\sfdefault}{bx}{n}

\usepackage{hyperref}
\hypersetup{hypertexnames=false}
\usepackage{url}
\usepackage[svgnames]{xcolor}
\usepackage{booktabs}
\usepackage{array}
\usepackage{threeparttable}
\usepackage{wrapfig}
\usepackage[most]{tcolorbox}
\usepackage{algorithm}
\usepackage{algpseudocode}
\usepackage{booktabs}
\usepackage{longtable}
\newtcbtheorem[
  auto counter
]{findingbox}{Finding}{
  enhanced,
  boxrule=0.1mm,
  colback=MediumTurquoise!10!white,
  colframe=DarkBlue!40!black,
  before skip=3pt,
  after skip=3pt,
  boxsep=1pt, 
  title={\textbf{Finding}~\thetcbcounter},
  toptitle=1pt,
  top=1pt,
  bottom=1pt,
  left=2pt,
  right=2pt
}{fnd}

\title{Multilinguality in Hybrid Attention LLMs}

\iclrfinalcopy

\author{
  \begin{tabular}{lllll}
    Lucas Bandarkar\thanks{Equal contribution; co-first authors.} & Junlin Hu\footnotemark[1] & Chenyuan Yang$^\dagger$ & Mohsen Fayyaz & Nanyun Peng \\
    \multicolumn{3}{r}{\normalfont University of California, Los Angeles} & \multicolumn{2}{c}{$^\dagger$\normalfont Fudan University}
  \end{tabular}%
}

\newcommand{\olmo}{OLMo}

\begin{document}

\maketitle
\fancyhead{} % Remove the ICLR publication header set by the conference style.

\begin{abstract}
In response to the growing demand for long sequences in agentic and reasoning use cases, many state-of-the-art LLMs combine multiple variants of attention to mitigate the quadratic complexity of traditional softmax attention.
These hybrid attention LLMs aim to balance the strengths and limitations of full attention and alternatives based on recurrence. This work presents a first study of how hybrid attention impacts the multilinguality of LLMs. 
Beyond the impact on long sequences in poorly tokenized languages, our study is motivated by the possibility that the inductive biases of the recurrent state alter linguistic processing.
Our interpretability analysis confirms this, showing that cross-lingual representations in hybrid models develop in patterns tied to the ordering of recurrent and full-attention layers.
Across diverse models, we notably observe a pronounced spike in cross-lingual alignment around the first full-attention layer.
These findings lead us to question the conventional ordering of attention layers.
% We run distillation experiments with multiple alternative layer arrangements and find that all fit multilingual data faster than the standard arrangement, some by over 60\%.
% In distillation experiments on multilingual data, all our alternative layer arrangements cleanly outperform the standard arrangement; some match its performance with only 40\% of tokens.
In distillation experiments on multilingual data, all alternative layer orderings outperform the standard throughout training, learning up to 2.5X faster.
These stark, replicable results prompt our theory that multilingual models would benefit from starting with a full-attention layer rather than recurrent layers.

\end{abstract}

\section{Introduction}

% - LLMs have started shifting to hybrid attention architectures \citep{kimi3}

% - chain-of-thought reasoning
% - agentic performance requires iterative, tool-calling, in-context data processing, etc

% -recent years has seen a rebirth of in research in recurrent/state-based architectures for LLMs
% - there remains no singular token-mixing mechanism that outperforms the rest across requirements, so developers have turned to hybrid attention architectures to mix and match the benefits of multiple

% As LLMs become practical for a wider range of use cases, their capability to work in many languages becomes increasingly important, whether that be for handlingmultilingual users or the large-scale processing of multilingual data \citep{omnilingualgaia2}.

% However, recurrent attention works fundamentally different from traditional attention, and it can be assumed the representational structure is very different. It remains an open question whether the large body of research on how LLMs handle multilinguality changes when dealing with these new forms.

The tasks we rely on LLMs for have become increasingly complex, demanding autonomous planning, tool-calling, large-scale data retrieval, and more.
These workflows accumulate massive contexts, making the $O(T^2)$ computational cost of full softmax attention (where $T$ is sequence length) a growing bottleneck.
Hybrid attention models address this pressure by combining full attention with recurrent mechanisms that scale linearly in $T$. Hybridization has become the leading paradigm among state-of-the-art open-weight models, including Kimi K3 and the Nemotron, Qwen, and MiniMax families \citep{kimi3,nemotron3,qwen35,minimax01}.

Its strengths extend beyond efficiency: combining recurrence and attention can yield greater expressivity than either mechanism alone, leading to significantly improved pretraining \citep{waleffe2024empiricalstudymamba,merrill2026olmo}.
These mechanisms have distinct inductive biases: recurrent layers maintain a compressed, evolving state, while full attention directly accesses all earlier token representations. 
These differences may affect how models process linguistic form and build internal abstractions. Yet how hybrid attention LLMs organize these processes amongst this heterogeneity remains understudied \citep{shen2026collapse,afendulev2026attention}. 
Understanding this organization can clarify the strengths and limitations of hybridization and guide future model design.

Multilinguality is both an essential model capability and a valuable lens on model internals. 
% Unoptimized multilingual tokenization makes sequence efficiency especially consequential in multilingual regimes.
% Cross-lingual comparisons also help probe the balance between low-level, language-specific processing and more abstract representations that support knowledge and reasoning.
% We therefore ask: \emph{how does hybridization reshape cross-lingual representations, and what can this reveal about the roles and ordering of attention layers?}
LLMs must serve users and process information across languages, a requirement that extends even to long-horizon autonomous agents deployed on real-world data \citep{omnilingualgaia2}. However, no prior work has studied how recurrent attention and hybridization impact multilingual processing.
Beyond being a major use case, cross-lingual data offers unique insight into model computations.
Comparing equivalent content across languages helps distinguish processing tied to surface form from more abstract representations shared across languages \citep{lee-etal-2025-multimodal,knowledgelocalization}.
%% Next sentence needs some adaptation
We therefore use multilinguality to study how hybrid LLMs organize computation.

In this work, we study four hybrid models alongside close non-hybrid counterparts. Because these pairs are not perfectly comparable, our analysis extends deeper than benchmark results.
In Section~\ref{sec:multilingual-tokenization}, we examine how token inflation makes long-sequence efficiency especially consequential for multilingual inputs, separating the ability to \emph{accommodate} longer sequences from the ability to retrieve their contents. Hybrid attention offers large efficiency gains on multilingual inputs, while its retrieval shortcomings actually lessen as contexts grow.

Then in Section~\ref{sec:cross-lingual-analysis}, we trace layer-wise representational patterns and their relationship to attention ordering. We show that in comparison to the smooth development of language-abstract representations exhibited in full attention models (and also in SWA-hybrid models), hybrid attention models exhibit much more abrupt changes around interleaved full attention layers. In particular, we identify a pronounced reorganization around the first occurrence of full-attention.
% add this space back for arxiv

Motivated by these patterns, in Section~\ref{sec:hybrid-distillation} we question the status quo of layer ordering adopted by major labs: periodically interleaving full-attention layers after recurrent layers. Rationale for these layer orderings is scarce, and multilinguality was likely not a primary consideration. 
We evaluate four alternative orderings by training hybrid models by distilling from a full-attention teacher LLMs. \emph{All} of them comfortably outperform the conventional periodic layout on multilingual data. The only characteristic shared by all five is that the first layer is full attention. And while pretraining from scratch is a very different setting, we formulate the theory that multilingual LLMs should start with full softmax attention.

% TODO: explicit contributions ?? or just do this in conclusion ?
% 1. evidence that inductive bias causes consistent shift in representations
% 2. numerous tidbits we find along the way (and friends?)
% 3. multilingual model design recommendations: use hybrid attention but start with full attention layer.
\section{Background and Related Work}

\subsection{Research in Attention Alternatives}

The attention mechanism \citep{attention} was originally introduced to address the long-sequence performance bottleneck of recurrent neural networks like LSTMs \citep{lstm}, which are constrained by the size of the state. However, transformer models \citep{vaswani2017attention} suffer from quadratic asymptotic complexity $O(T^2)$ with respect to sequence length $T$ because of the $\displaystyle \mQ\mK^T$ operation. Popular solutions proposed to achieve linear complexity $O(T)$ were to use low-rank matrix approximations \citep{linformer} or sliding-windows (SWA) \citep{longformer}. Notably, the FlashAttention \citep{flashattention} implementation of full attention achieves linear complexity of peak-memory consumption, even if computational complexity remains quadratic.

However, \citet{pmlr-v119-katharopoulos20a} showed that without the softmax operator, self-attention could be reduced via matrix associativity to a linear-complexity recurrence relation. This prompted a line of research that seeks to reach attention-level expressivity and parallelization while maintaining linear complexity, including selective state-space models \citep{gu2024mamba} and the Delta Rule \citep{yang2024parallelizing}. More recent recurrent attention variants, such as Gated DeltaNet \citep{gdn} and Kimi Delta Attention \citep{kimilinear}, improve the in-context learning ability of state-space models by maintaining an updatable associative key-value memory \citep{qwen3next}.

% - \citep{cabannes2026sparsedeltamemoryscaling} scales GDN state using sparsity

\subsection{Hybrid Attention LLMs}

Modern LLMs increasingly combine these recurrent mechanisms with occasional full-attention layers to exploit their complementary strengths \citep{mehta2023long,fu2023hungry,lenz2025jamba,minimax01}. Recurrent layers offer efficient state tracking over long sequences, while full attention preserves precise content-based retrieval that is difficult to compress into a fixed-size state \citep{qiao2026rethinkingroleefficientattention}. Beyond reducing inference costs, this combination can be more expressive and scale more efficiently during pretraining than either mechanism alone \citep{merrill2026olmo}. Hybridization can occur within a single attention block \citep{falconh1,du-etal-2026-native}, but more commonly, the two attention architectures are interleaved in subsequent decoder layers.

\paragraph{Terminology} In this work, we refer to traditional softmax attention as ``full'' attention and linear-complexity recurrent alternatives generally as ``recurrent'' attention. We categorize SWA separately, since it is linear-complexity, sparse local attention. For simplicity, we use ``attention'' broadly to include components such as Mamba-2, even if \emph{sequence mixers} is the more general term.

\subsection{Multilinguality in LLMs} \label{multilingual_works}

Recent multilingual interpretability research has notably found that LLMs operate in a language-shared representations in the middle layers \citep{kojima-etal-2024-multilingual,zhao2024multilingualism,wu2025the,dumas-etal-2025-separating}. This follows language-specific detokenization \citep{lad2025remarkable} in early layers and precedes language-specific generation in the final layers. 
Analogous to this work, \citet{bandarkar2026multilingual} studies how the adoption of sparse mixture-of-experts layers impacts multilinguality in LLMs by designing metrics to analyze cross-lingual alignment in the MoE layer.
The attention mechanism, in contrast to the feed-forward network, provides access to previous tokens, making it essential for language-specific, low-level linguistic processing.
Prior work links attention heads to syntactic processing and multilingual performance \citep{voita-etal-2019-analyzing,ma-etal-2021-contributions,zhang2025samebutdifferent,liu2026focusing}.
Cross-lingual fine-tuning research finds attention updates to be much more essential than FFN updates \citep{bandarkar2025layerswapping,bandarkar-peng-2025-unreasonable,tran-etal-2026-disentangling}.
These findings generally suggest attention plays a disproportionate role in model multilinguality.

\section{Models} \label{performance}

For our investigation, we analyze four hybrid-attention models with reasonably close counterparts for comparison, listed in Table~\ref{tab:model-comparison}. While options were limited, these four pairs are diverse in size, sparsity, multilinguality, and more. Three periodically interleave full attention \emph{after} 3 or 4 recurrent attention layers, while Granite-4.0-H places full attention in layers 4, 14, 24, and 34.

\begin{table}[h]
\caption{Hybrid-attention models studied in this work and their closest non-hybrid counterparts. In the interest of space, we provide further design details and citations for all in Appendix~\ref{models_details}. The much larger Ling-2.6-flash and SWA-hybrid Tiny Aya are also studied partially.}
\label{tab:model-comparison}
\centering
\small
\setlength{\tabcolsep}{3pt}
\renewcommand{\arraystretch}{1.2}
\begin{threeparttable}
\begin{tabular}{@{}p{0.15\linewidth}|p{0.17\linewidth}p{0.05\linewidth}p{0.09\linewidth}|p{0.12\linewidth}p{0.35\linewidth}@{}}
\toprule
\textbf{Hybrid-Attn Model} & \textbf{Recurrent \hspace{3em} Variant} & \textbf{Ratio} & \textbf{Full Attn Variant} & \textbf{Comparable Non-Hybrid} & \textbf{Relationship} \\
\midrule
Qwen3.5-35B-A3B & Gated DeltaNet & 3:1 & Gated GQA & Qwen3-30B-A3B & Unknown, same architecture outside attention\\
\midrule
\olmo-Hybrid & Gated DeltaNet w/ neg. eigenvalues & 3:1 & GQA & \olmo-3-7B\tnote{a} & Exact same data recipe, trainings from scratch controlled for comparability.\\
\midrule
Ring-mini-linear-2.0 & Lightning Attn 2 & 4:1 & MLA & Ring-mini-2.0 & Same base model, with hybrid conversion via distillation happening before comparable post-trainings \\
\midrule
Granite-4.0-h-micro & Mamba-2 & 9:1\tnote{b} & GQA & Granite-4.0-Micro & Same data recipe, each trained from scratch, unknown how controlled  \\
\hline
\hline
Ling-2.6-flash & Lightning Attn 2 & 7:1 & MLA & --- & --- \\
\midrule
Tiny Aya Global & SWA & 3:1 & GQA & --- & --- \\
\bottomrule
\end{tabular}
\begin{tablenotes}[flushleft]
\footnotesize
\item[a] \olmo-3 is technically a ``hybrid'' model since it interleaves SWA layers at a 3:1 ratio, just like Tiny Aya.
\item[b] Granite-4.0-h-micro uses a non-regular interleaving of four full-attention layers among 40 total; see the \href{https://huggingface.co/ibm-granite/granite-4.0-h-micro/blob/main/config.json}{config}.
\end{tablenotes}
\end{threeparttable}
\end{table}

The recurrent variants in these models differ primarily in how they update their recurrent states.
Lightning Attention-2 in Ring-linear accumulates key-value associations with input-\emph{independent} decay. Qwen3.5's Gated DeltaNet instead combines input-dependent forget gates with delta-rule updates.
Granite uses Mamba-2, a selective state-space model that also uses forget gates, but with additive updates.
\olmo-Hybrid uses a close variant of Gated DeltaNet \citep{grazzi2025unlocking}.

None of these pairs enable a flawless comparison: even though the \olmo~pair's controlled experiments are designed for comparing, its explicit exclusion of multilingual training data makes it the \emph{least} useful for our purposes. 
We initially compare the models on a few standard multilingual benchmarks, but find no consistent evidence of a cross-lingual transfer advantage (details in Appendix~\ref{benchmarks}). Across the four pairs, hybrid models are generally stronger overall and, except for the jointly released Granite 4.0 models, were released later than their non-hybrid counterparts, complicating 1-to-1 comparisons. With matched training conditions, \olmo-Hybrid consistently performs slightly better than \olmo-3. However, when examining the multilingual \emph{gap} relative to English as an indicator of cross-lingual transfer, the relative advantage varies from benchmark to benchmark. Thus, we extend our analysis past simple evaluation.

\section{Multilingual Tokenization}
\label{sec:multilingual-tokenization}

Long-sequence efficiency is particularly relevant to multilinguality because of typically poor tokenization in many languages.
Even a large-vocab tokenizer like Qwen3.5 creates sequences that are 2.6$\times$ in Yoruba than English (estimated on parallel data). The English-centric \olmo~explodes contexts 10$\times$ in some non-Latin-script languages. So here, linear-complexity attention is critical.

We evaluate the long-context recall cost in English versus other languages using five OneRULER needle-in-a-haystack tasks \citep{oneruler} (excluding common-word extraction), with results in Table~\ref{oneruler}. On a single A6000 GPU with 48GB VRAM, our evaluations encountered out-of-memory (OOM) failures substantially less often for hybrid models, as expected.
At 128K, Qwen3.5 completed 25/40 language-task evaluations, compared with only one for Qwen3.
For Granite models, the 64K evaluations on the hybrid variant took considerably shorter: 13.5 hours versus 61.5.

% Mohsen: is this only because of hybrid speed? or is the non hybrid generating more tokens too?
% Lucas response: good question... i dont know if the model has reasoning on this task

\begin{findingbox}{}{}
Hybrid attention offers a major memory advantage for token-inflated multilingual sequences, while its performance cost actually diminishes as context length grows.
\end{findingbox}
% \vspace{-1em}
% Does this memory advantage come at a retrieval cost? \footnote{We omit the \olmo~pair pending an evaluation audit: \olmo-3 is near zero even in English, which multilingual tokenization alone cannot explain.}

\begin{table}[h]
\centering
\small
\setlength{\tabcolsep}{5pt}
\caption{OneRULER accuracy $\uparrow$ (\%) on needle-in-a-haystack (NIAH) tasks at different context sizes.
Note that Qwen3.5 is 9 months newer, bigger, and better than Qwen3.}
\label{oneruler}
\begin{tabular}{@{}l|rrr|rrr@{}}
\toprule
 & \multicolumn{3}{c}{English} & \multicolumn{3}{c}{Non-Eng (10 langs)} \\
% \cline{2-4}\cline{5-7}
Model & 8K & 32K & 64K & 8K & 32K & 64K \\
\midrule
Qwen3-30B-A3B & 99.2 & 98.0 & 97.6 & 98.2 & 95.3 & 92.0 \\
Qwen3.5-35B-A3B & 100.0 & 100.0 & 99.6 & 96.9 & 95.6 & 93.8 \\
\midrule
Granite-4.0-Micro & 80.4 & 65.6 & 47.2 & 71.2 & 47.9 & 38.1 \\
Granite-4.0-H-Micro & 75.2 & 68.0 & 61.6 & 52.1 & 39.5 & 34.9 \\
\bottomrule
\end{tabular}
\end{table}

OneRULER notably modifies the ``haystack'' based on each language and tokenizer so that context length remains fixed (at 8K/32K/etc.). As a result, non-English languages contain less underlying content at the same length during evaluation. We hypothesized that the lower information density of poorly tokenized languages would make the retrieval gap between hybrid and full attention smaller than in English. Instead, the hybrid deficit is initially larger outside English. And surprisingly, this deficit \emph{narrows} as context length grows in both English and non-English languages. Focusing on Granite, the more comparable pair, Granite-H trails Granite substantially on non-English inputs at the shortest context.
\begin{wrapfigure}[18]{r}{0.35\linewidth}
  \vspace{-0.5\baselineskip}
  \centering
  \includegraphics[width=\linewidth]{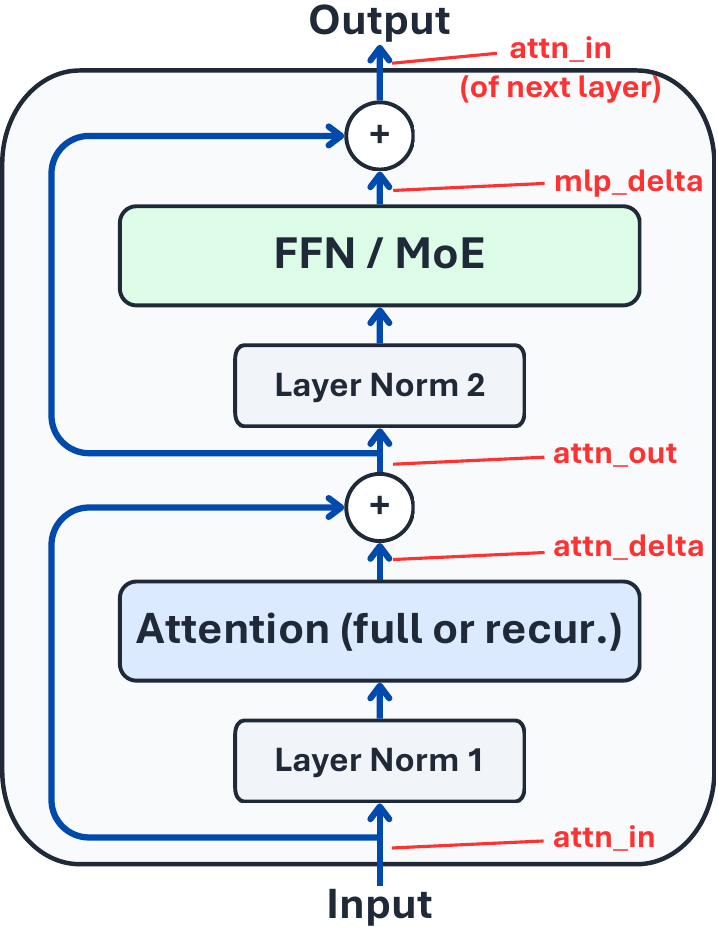}
  \caption{Formalization of the decoder layer structure applicable for all decoder layers across models.}
  \label{diagram}
%   \vspace{-2\baselineskip}
\end{wrapfigure}
Poor performance on the lowest-resource language, Sesotho, contributes the most to this gap. In English, however, Granite-H overtakes Granite at longer contexts, and the non-English gap likewise narrows, even if Granite-H still trails at 64K. The Qwen pair shows a similar trend, although with a much smaller difference. Overall, the performance cost is smallest at the long contexts where the memory and runtime advantages of hybrid attention matter most.

% Surprisingly, at 8K Granite-H has a larger average gap in non-English languages, driven in large part by the lowest resource language in OneRULER, Sesotho.
% However, hybrid models demonstrate similar degradation in non-English languages to English, even though the ``haystack'' contains less information per token. Non-hybrid models seem

% To distinguish multilingual behavior from Qwen3.5's broader improvements, we compare each model's non-English scores with its English scores. Despite similar or higher raw accuracy, Qwen3.5's English-to-non-English gap is larger than Qwen3's by 2.0, 1.7, and 0.2 percentage points at 8K, 32K, and 64K, respectively. Its gains therefore do not fully carry over beyond English. This relative disadvantage shrinks with length across all five tasks, but grows when no-needle detection is excluded. Granite shows the opposite pattern: its hybrid has lower absolute accuracy but a smaller gap relative to English.

\section{Visualizing Representations}
\label{sec:cross-lingual-analysis}

\subsection{Cross-Lingual Alignment Metric: SoftCKA}

We investigate numerous approaches to understanding how cross-lingual representations evolve through the model layers.
Because of the diversity of attention implementations, our analysis treats the attention block as a black box. We focus on the hidden state entering the decoder layer (labeled as \verb+attn_in+ in Figure~\ref{diagram}) and after the attention block; specifically, after the residual stream connection (\verb+attn_out+). A major obstacle to measuring cross-lingual alignment from these states is that tokenization and linguistic structure prevent a one-to-one mapping of tokens across texts. Typical measurements instead mean-pool over a sequence or use only its last token. We attempt to mitigate the heavy information loss of such metrics with a metric that \emph{softly} aligns tokens.
Our metric, SoftCKA, is based on centered kernel alignment (CKA) \citep{pmlr-v97-kornblith19a}, which compares neural network hidden states pair-wise through the geometric similarity of the representation space instead of absolute values.
Let $\mX_1,\mX_2$ be the sequences of hidden states at a particular point in the model on a pair of parallel (i.e. translated) sentences. That is, $\mX_1 \in \mathbb{R}^{T_1 \times d}$ and $\mX_2 \in \mathbb{R}^{T_2 \times d}$, where $T_1,T_2$ are the sequence lengths in languages 1,2 respectively.

We first compute cross- and within-language RBF kernel matrices $\mK_{12}, \mK_{11}, \mK_{22}$.
Each entry $\mK_{12}[i,j]$ measures the similarity between token $i$ in language 1 and token $j$ in language 2.
These similarity scores provide a form of ``soft matching'' without an explicit token assignment.
We then row- and column-center these matrices before taking the Frobenius norm, so that $h_{12} = \lVert \widetilde{\mK}_{12} \rVert_F^2$, and the same for $h_{11}, h_{22}$.
The within-language quantities $h_{11}$ and $h_{22}$ are
proportional to the Hilbert-Schmidt independence criterion
(HSIC) evaluated between each representation and itself, used in CKA's normalization. Meanwhile, $h_{12}$ is analogous to the HSIC term comparing two representations in standard CKA, without requiring
explicit token pairing.

\begin{wrapfigure}[12]{r}{0.55\linewidth}
    \vspace{-1\baselineskip}
    \centering
    \includegraphics[width=\linewidth]{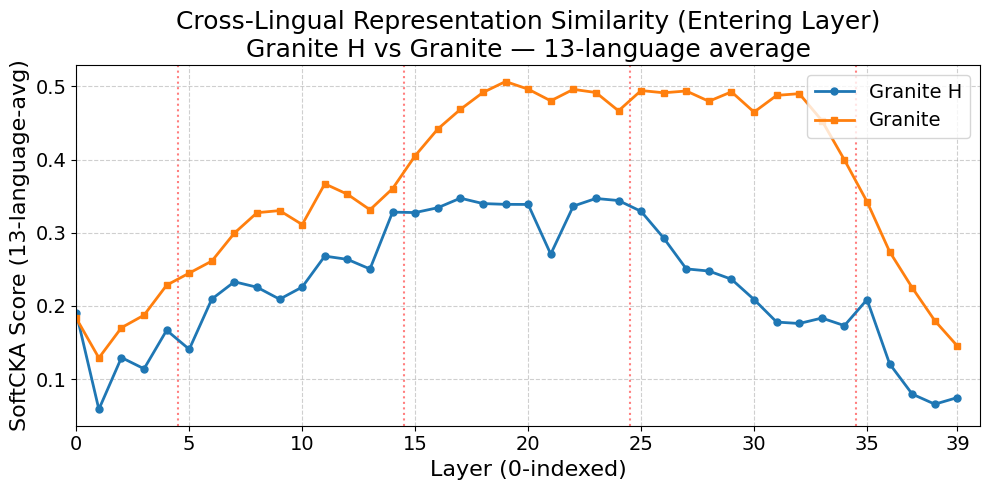}
    \caption{SoftCKA scores for Granite models, with hidden states taken at the beginning of each decoder layer (\texttt{attn\_in}). Red dotted lines mark where full attention layers are in Granite H.}
    \label{fig:granite-13lang}
\end{wrapfigure}

\begingroup
% \setlength{\abovedisplayskip}{4pt}
% \setlength{\belowdisplayskip}{4pt}
% % \setlength{\abovedisplayshortskip}{2pt}
% \setlength{\belowdisplayshortskip}{2pt}
% \vspace{-1\baselineskip}
\begin{equation}
    \operatorname{SoftCKA}(\mathcal{D})
    =
    \frac{
        \sum_{s \in \mathcal{D}} h_{12}^{(s)}
    }{
        \sqrt{
            \left(\sum_{s \in \mathcal{D}} h_{11}^{(s)}\right)
            \left(\sum_{s \in \mathcal{D}} h_{22}^{(s)}\right)
        }
    }.
    \label{eq:softcka}
\end{equation}
\endgroup
We aggregate these quantities over a corpus $\mathcal{D}$, which is a set of sentences parallel in languages 1,2.
Centering, corpus weighting, and implementation details are provided in Appendix~\ref{app:softcka}.
%add this space back for arxiv
This score, a modification of CKA, normalizes cross-language kernel variation by the corresponding within-language quantities.
We find this metric for cross-lingual alignment much less noisy than alternatives we explore.
On non-hybrid LLMs, SoftCKA scores smoothly rise toward the middle and decline in the final layers (See Figures~\ref{fig:granite-13lang},\ref{fig:vis-jagged},\ref{tinyaya_layers}), consistent with multilingual interpretability studies that use very
different approaches (See Section~\ref{multilingual_works}).
This provides supporting evidence for our interpretation of the metric, although smoothness cannot prove its faithfulness.

In this analysis, we use parallel data from FLoRes \citep{nllb2022} and calculate SoftCKA alignment between English and a highly diverse subset of 13 languages that cover different language families, scripts, and resource-levels: French, Chinese, Hindi, Thai, Farsi, Bengali, Serbia, Darija Arabic, Modern Standard Arabic, Lithuanian, Bambara, Orya, Assamese.

\subsection{Observational Findings}
\label{sec:observation}
We start by using SoftCKA to measure representational alignment to English.

\begin{findingbox}{}{}
Representations are less aligned across languages in hybrid attention LLMs.
\end{findingbox}

As shown in Figure~\ref{fig:granite-13lang}, Granite-H builds significantly less aligned representations, particularly in the middle layers. The figure displays a multi-language average, but this is individually the case for all 13 languages. For Qwen models, Qwen3.5 has less cross-lingual alignment in 10/13 languages. For Ring, Ring-linear has lower alignment in 11/13 languages. For \olmo~models it is mixed.
And while multilingual NLP literature continuously finds that lower cross-lingual alignment is worse for cross-lingual transfer \citep{cao2020multilingual,deshpande-etal-2022-bert,gaschi-etal-2023-exploring,lim2025language,bandarkar2026multilingual}, we do not claim that this lower alignment is bad. It is conceivable, even if unlikely, that recurrent attention simply requires lower alignment for equivalent transfer.

\begin{findingbox}{}{}

Hybrid LLMs specifically organize multilingual processing based on the layer ordering, with abrupt changes in cross-lingual metrics around full attention layers.
\end{findingbox}

\begin{figure}[h]
    \centering
    \includegraphics[width=\linewidth]{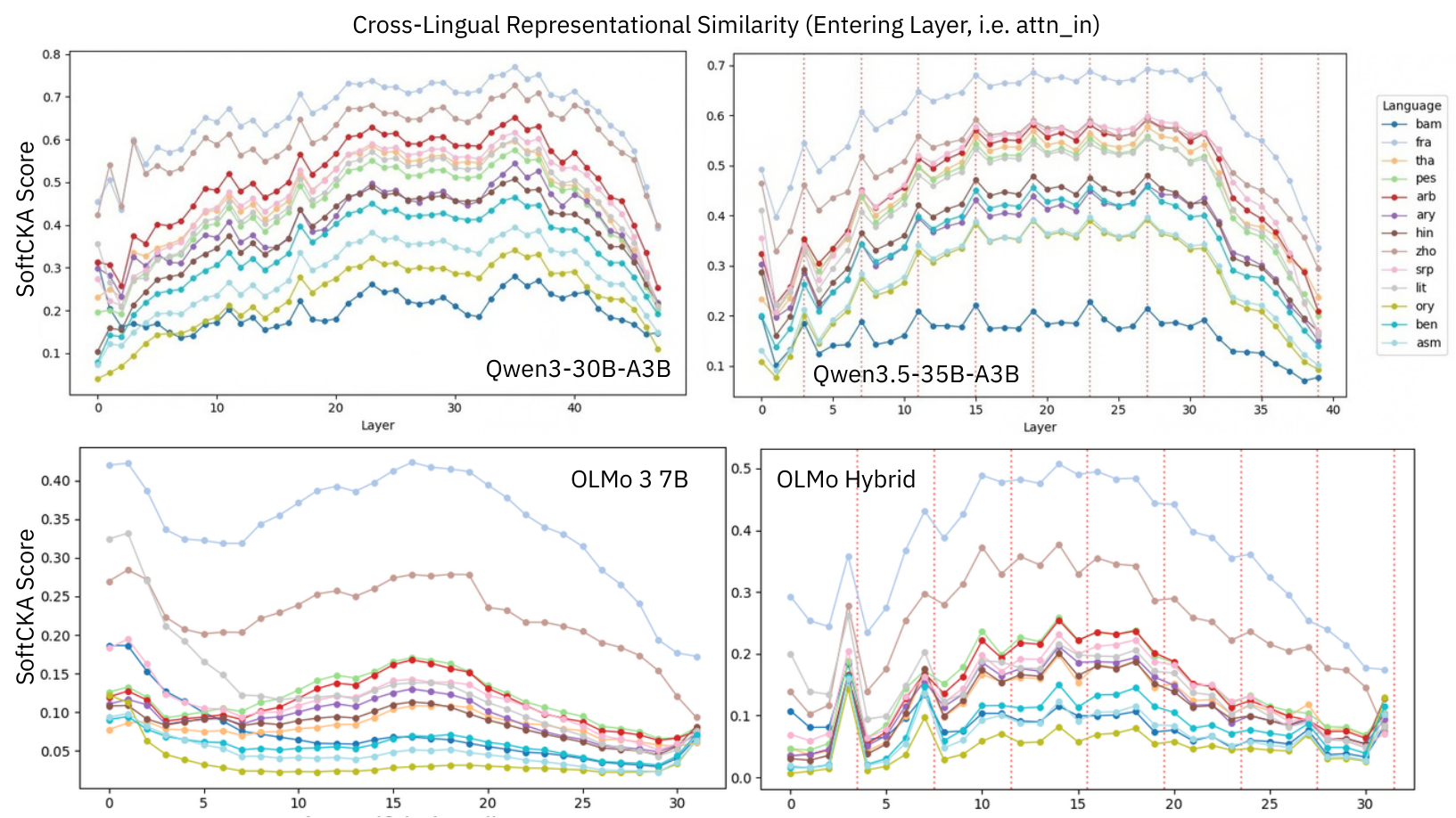}
    \caption{Visualization of the jagged curves in the hybrid models (right), which seem to reflect the heterogeneous attention layout, compared with the smoother curves in non-hybrid counterparts(left).}
    \label{fig:vis-jagged}
\end{figure}

The most notable consistent pattern we identify is that SoftCKA reveals homogeneous LLMs smoothly build shared representations and then undo them before generation while hybrid LLMs experience many abrupt changes every time a full attention layer comes around. This difference is consistent across all 4 model pairs, 2 of which are shown in Figure~\ref{fig:vis-jagged}.
Specifically, in agreement with Finding 2, the representations get more aligned before each full attention block.

\begin{findingbox}{}{} \label{xl_event}
A major cross-lingual event occurs at first full attention layer
\end{findingbox}

\begin{wrapfigure}[16]{l}{0.5\linewidth}
    \vspace{-1\baselineskip}
    \centering
    \includegraphics[width=\linewidth]{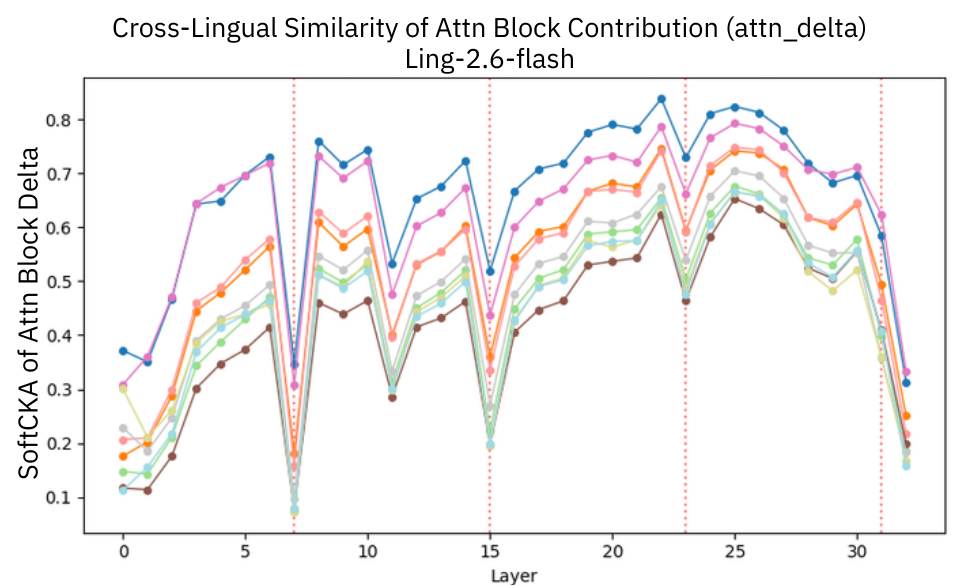}
    \caption{SoftCKA scores of the attention block \emph{delta}. Curiously, the major ``event'' at the first full attention layer is opposite to other models.}
    \label{fig:ling26flash-delta}
\end{wrapfigure}

We also identify that the most abrupt spike happens at the \emph{first} full attention layer. This is consistent across all hybrid models in the four pairs and additionally Ling-2.6-flash.

As a secondary metric, we also calculate the SoftCKA of the output of the attention block, \verb+attn_delta+, which measures the cross-lingual similarity of a block's contribution to the residual stream\footnote{Note: with the residual connection, the block output \texttt{attn\_delta} is just \texttt{attn\_out} - \texttt{attn\_in}}. This metric most dramatically shows that for Qwen3.5, \olmo-Hybrid, and Granite-H, the hidden state transformation in the first full attention block is very aligned across languages relative to layers around it. Very curiously, for Ring and Ling-2.6-flash, there is a large \emph{negative} spike instead, meaning the full attention block contribution is less similar across languages (Figure~\ref{fig:ling26flash-delta}). Because of this inconsistency, we resort to using the vague term ``event''. Regardless, it is clear that the LLMs are markedly organizing around this layer. We also find that the preceding MLP and attention blocks progressively prepare inputs for this, see Appendix~\ref{softcka_delta}.

% A third--completely different--metric, the mixture-of-experts (MoE) routing alignment metric from \citet{bandarkar2026multilingual}, also identifies a first attention layer phenomena on MoE models Qwen and Ring. In non-hybrid MoEs, routing divergence starts decreasing immediately, while in the hybrid versions, it does not begin to decrease until after the first full-attention layer, see Appendix~\ref{moe_vis}.

\begin{findingbox}{}{}
Hybridization with SWA does \emph{not} lead to such a representational re-organization
\end{findingbox}

Instead of recurrent attention hybrids, many modern LLMs address long-context efficiency by using sliding window attention (SWA) layers. We repeat the above visualizations for the SWA-hybrid models Tiny Aya Global and \olmo-3. The abrupt changes occurring at full-attention layers, particularly at the first, are absent; see Appendix~\ref{swa_hybrid}. Across various metrics, we find no noticeable pattern different than full-attention LLMs. We conclude that the inductive biases of SWA are not different enough to prompt the reorganization observed in LLMs that interleave recurrent layers.

\begin{findingbox}{}{} \label{attn_distance}
Irrespective of language, token attribution shows full attention attends more locally in hybrid models than typical attention layers in non-hybrid models.
\end{findingbox}

We study how global or local attention is across different layers. Full attention, by definition, produces a masked matrix $\mA = \displaystyle \mQ\mK^T / \sqrt{d_k}$ that can serve as an explanation of what previous tokens influenced the current. There are methods proposed to re-compose this matrix in recurrent attention via unrolling \citep{guo2026loglinear}, but comparing cross-architecture is unreliable. As a result, we simply compare the full attention layers and find that the few full attention layers in hybrid models devote a larger proportion of their attention to the most recent tokens (in comparison to full attention layers in non-hybrid models). And even though token attribution from the attention matrix alone may not be entirely faithful \citep{modarressi-etal-2023-decompx, modarressi-etal-2022-globenc}, we find this to be very consistent across models and all languages, and not layer-dependent. This could suggest hybrid models reserve local tasks to full attention, which is not necessarily a contradiction to the established notion that full attention takes care of long-range exact retrieval \citep{afendulev2026attention}.

Because of these findings, primarily 3, 4, and 6, we hypothesize that another layer ordering may better facilitate the abstraction of multilingual inputs into language-independent representations.

\section{Hybrid Distillation Experiments}
\label{sec:hybrid-distillation}

\subsection{Related Work on Full Attention Placement}
How and where to interleave attention-layer types remains poorly understood: frontier labs almost never disclose the rationale for their architectural choices, beyond noting that the \emph{ratio} is empirically optimized \citep{systematicanalysis,qwen3next,ring}. Periodic interleaving has been justified by the notion that it allows the model regular, exact access to previous tokens \citep{dao2024transformers,ren2025samba}, though this was on Mamba hybrids specifically. Separately, \citet{kimilinear} argues that periodic interleaving simplifies KV-cache management.

Within periodic blocks, the status quo seems to be recurrent attention before full (N:1 instead of 1:N). \citet{waleffe2024empiricalstudymamba} notes that starting with Mamba means later layers don't need positional embeddings. \citet{gdn} tests numerous orderings and ends up placing full attention last in their repeating block.  Meanwhile, broader experiments from distillation works result in diverse conclusions \citep{yang2025zebrallama,li2026distilling,gu2025jet, xia2026distillthenreplaceefficienttaskspecifichybrid}. Given this limited research and the fact that it is quite unlikely multilingual considerations were taken into account, we experiment with multilingual data in controlled layer placements.

\subsection{Experimental Setup}

% Distillation is a popular way to train hybrid models, such as with Ring-linear. 
Hybrid models can be pretrained from scratch, but are often distilled from a homogeneous teacher model for resource-efficiency \citep{wang2024mambainllama,li2026distilling,xia2026distillthenreplaceefficienttaskspecifichybrid}, such as Ring-Linear.
We run experiments distilling from two small full attention models, Qwen3-4B \citep{yang2025qwen3technicalreport} and (secondarily) Granite-4.1-3B \citep{granite2026} based on the RADLADS \citep{goldstein2026radladsrapidattentiondistillation} and HALO \citep{chen2026hybridlinearattentionright} methods. For further efficiency, we mostly adopt the student initialization method from HALO. We start by converting select layers to Gated DeltaNet,
% according to the placement schemas,
inheriting the query, key, value, and output projections from the teacher's full attention of the corresponding layers. Parameters without a counterpart in the teacher are initialized randomly. We then train the whole student to match the frozen teacher's next-token distribution under KL-divergence $D_{\mathrm{KL}}$. Unlike RADLADS and HALO, we skip the preliminary stage that trains each converted layer to reproduce the hidden states of the attention layer it replaces as this would undermine our comparisons.
% That stage asks every recurrent layer to take over the role of the full-attention layer before it, whereas Section~\ref{sec:observation} suggests that the two layer types play different roles;
Training only on the output distribution allows the student of hybrid attention structure to organize multilingual processing according to its layer ordering.
%, which is the effect we want to maintain. Within our compute budget, 
We otherwise follow the training recipes and hyperparameters of RADLADS and HALO closely (with minor adjustments for a $D_{\mathrm{KL}}$-only setup) as we lack the resources for tuning. They are detailed in  Appendix~\ref{app:hyperparameter}.

\paragraph{Data and Metrics} The training data is the highly multilingual, specifically FineWeb2 \citep{penedo2024finewebdatasetsdecantingweb}. We compose a subsample with 30\% English as anchor and the rest a mix of 26 languages that are diverse in families, scripts, and resource-level. We set aside 1000 documents for validation and 1000 for test set in each language (52K total). See full data mix details in Appendix~\ref{app:data}. We limit training runs to 1B tokens and report two main metrics: $D_{\mathrm{KL}}$ to the teacher and cross-entropy loss $\mathcal{L}_{\mathrm{LM}}$. In particular, we discuss the $\mathcal{L}_{\mathrm{LM}}$ \emph{gap} from the teacher's $\mathcal{L}_{\mathrm{LM}}$, which is, in theory, the upper bound of the student's modeling performance.

\begin{figure}[H]
    \centering
    \includegraphics[width=0.85\textwidth]{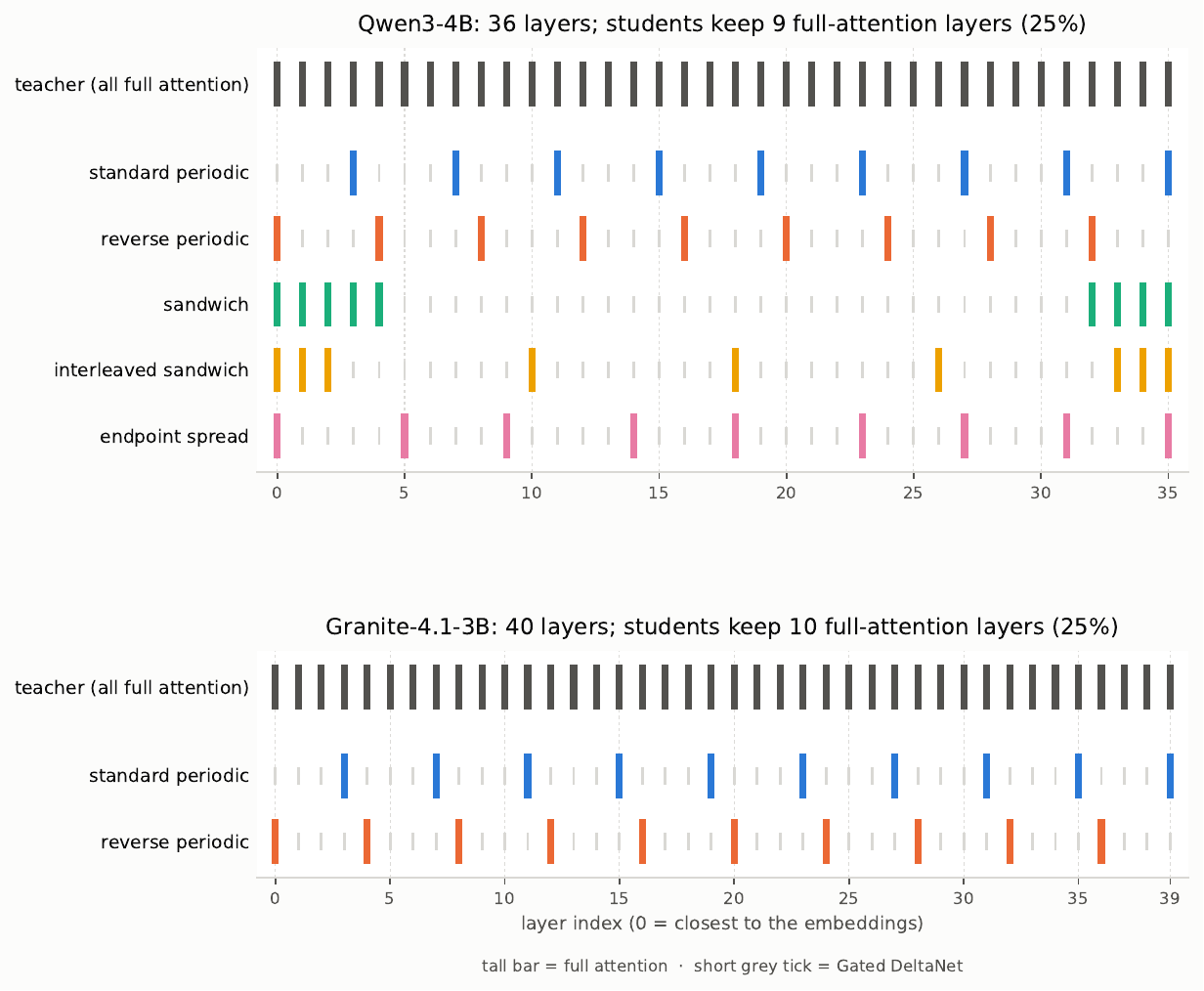}
    \caption{\textbf{Full attention placement schemas} at a fixed 25\% budget to maintain the 3:1 ratio between recurrent attention and full attention
    (9/36 layers for Qwen3-4B, 10/40 for Granite-4.1-3B). Tall bars = full
    attention, short grey ticks = Gated DeltaNet; layer 0 is closest to the
    embeddings. Students differ only in placement, not in how many full-attention
    layers they keep.}
    \label{fig:distillation_layouts}
\end{figure}

\paragraph{Experimental Layer Placements} We fix the ratio of full attention layers to be 25\% (9/36 layers in Qwen3-4B's and 10/40 for Granite-4.1-3B) and vary only where these go. We provide a convenient visualization of the different placements in Figure~\ref{fig:distillation_layouts} which we verbalize here.
We term the baseline layout, in which full attention closes each block of four layers, \textit{standard periodic}.
% \subsection{Alternative Full-Attention Placements}
\textit{Reverse periodic} is the smallest-change variant to it: full attention opens each block instead, which moves every full-attention layer three positions earlier and keeps the general block-interleaving pattern the same.
The other three placements are non-periodic. \textit{Sandwich}, the most extreme, clusters all full-attention layers at the first and last layers. \textit{Interleaved sandwich} keeps two-thirds of them at the ends and spread the rest periodically in the middle, and \textit{endpoint spread} places full attention in the first and last layers and spaces the remaining layers evenly in between.
% Because \textit{reverse periodic} keeps both the number and the spacing of full-attention layers, it is the most controlled comparison with \textit{standard periodic}, so we use this pair for the replicate run and for the second teacher.
The four alternatives differ from one another in many ways, but unlike \textit{standard periodic}, all of them use full attention in the first layer.

% \subsection{Distillation Training}
% \label{sec:distill-training}

% \paragraph{Method.} We build each student from its teacher by copying all of its weights and converting every layer outside the full-attention set to Gated DeltaNet, initializing the query, key, value, and output projections from the teacher's attention \citep{goldstein2026radladsrapidattentiondistillation,chen2026hybridlinearattentionright}; parameters without a counterpart in the teacher are initialized randomly. We then train the whole student to match the frozen teacher's next-token distribution under forward KL divergence. Unlike RADLADS and HALO, we skip the preliminary stage that trains each converted layer to reproduce the hidden states of the attention layer it replaces. That stage asks every recurrent layer to take over the role of the full-attention layer before it, whereas Section~\ref{sec:observation} suggests that the two layer types play different roles; training only on the output distribution allows the student divide the work according to its placement, which is the effect we want to maintain. Within our compute budget, we otherwise follow the training recipes of RADLADS and HALO, with small adjustments for a KL-only setup, and use the same hyperparameters for every placement without specific tuning(Appendix Table~\ref{tab:distill-hparams}).

% \paragraph{Confirming the placement for follow-up runs.}
% \subsection{Initial Qwen3-4B Runs}
\subsection{Results}

\begin{findingbox}{}{}
In our distillation experiments on multilingual data, \emph{all} our diverse alternative layer arrangements that start with a full attention layer significantly improve upon the standard layer ordering.
\end{findingbox}

\begin{figure}[!b]
    \centering
    \begin{minipage}[t]{0.49\linewidth}
        \centering
        \includegraphics[width=\linewidth]{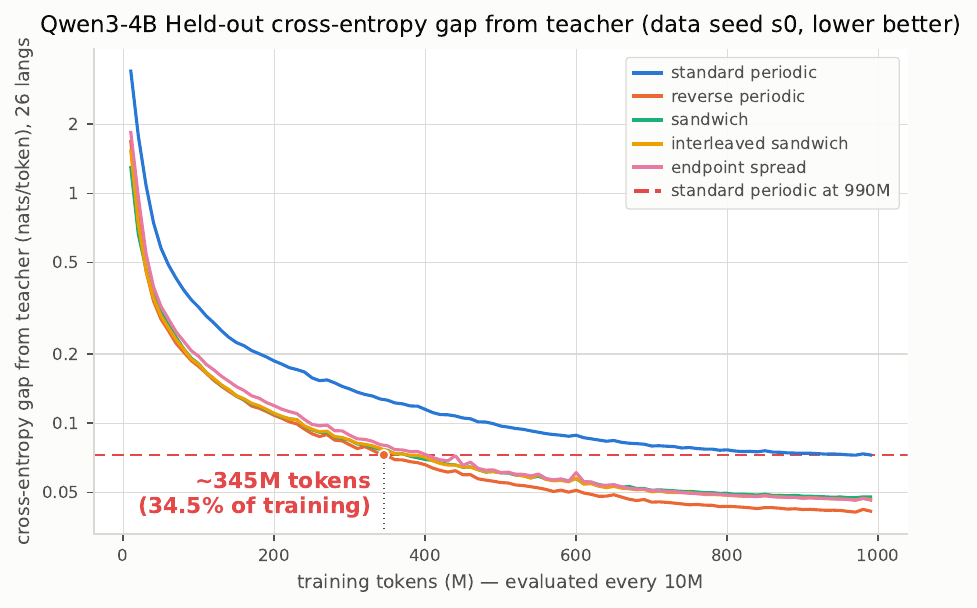}
        \small (a) Qwen3-4B student model (data seed s0)
    \end{minipage}
    \hfill
    \begin{minipage}[t]{0.49\linewidth}
        \centering
        \includegraphics[width=\linewidth]{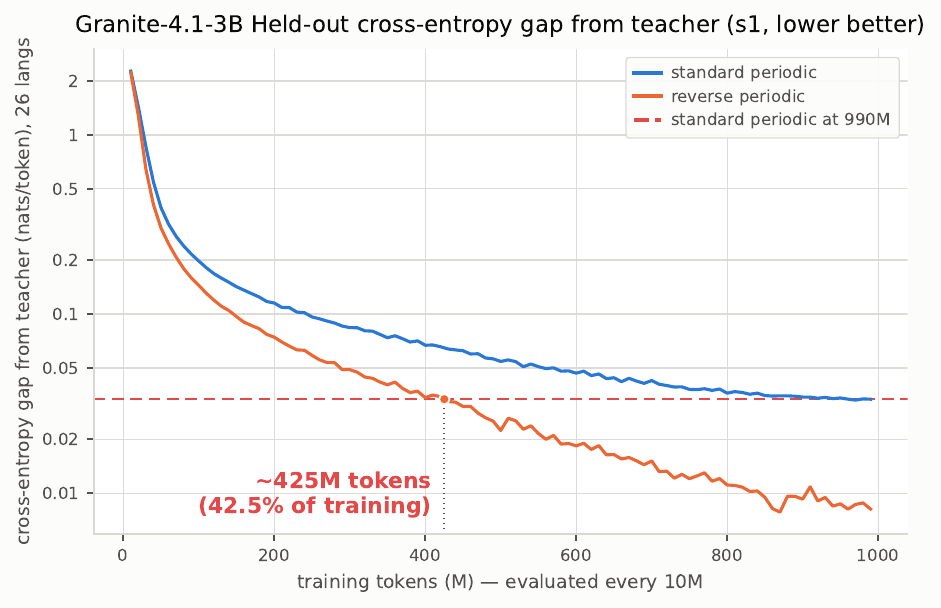}
        \small (b) Granite-4.1-3B student model (data seed s1)
    \end{minipage}
    \caption{Held-out $\mathcal{L}_{\mathrm{LM}}$ gap from the teacher (nats/token, mean over 26 languages, validation every 10M tokens). (a) Qwen3-4B, all five placements, data seed 0. (b) Granite-4.1-3B, \textit{standard periodic} and \textit{reverse periodic}, data seed 1. The red dashed line marks \textit{standard periodic}'s value at the last checkpoint (0.073 nats in (a) and 0.033 in (b) at 990M tokens). In (a), all alternatives reach it within $\sim40\%$ of training, with \textit{reverse periodic} at 34.5\%. In (b), \textit{reverse periodic} reaches it at 42.5\%.}
    \label{fig:distillation-eval-curves}
\end{figure}

We first distill Qwen3-4B into all five placements, including the baseline. 
We only budget 1B-token runs, and naturally, no run has converged by then.
% : held-out KL and cross-entropy are still falling for every placement over the last quarter of training, so multilingual training on larger scale would likely improve all students further.
But 1B tokens is plenty to see that every alternative learns much faster. On validation set, their lead over \textit{standard periodic} appears within the first eval step and holds at every later step. As displayed in Figure~\ref{fig:distillation-eval-curves}a, each reaches \textit{standard periodic}'s final cross-entropy value with just 40\% of tokens. Concretely, they have lower $\mathcal{L}_{\mathrm{LM}}$ and lower $D_{\mathrm{KL}}$ on train, valid, and test sets and in nearly every language. Within the alternatives:
On $D_{\mathrm{KL}}$, \textit{sandwich} and \textit{reverse periodic} are close: \textit{sandwich} has the lower average, but each is better in about half the languages. On cross-entropy, \textit{reverse periodic} is the clear leader: its mean gap to the teacher is 10\% smaller than any alternative and the lowest in 16/26 languages. See Table~\ref{tab:distill-main}.

% \textit{sandwich} at 375M (37.5\%), \textit{interleaved sandwich} at 378M (37.8\%), and \textit{endpoint spread} at 402M (40.2\%). In (b), \textit{reverse periodic} reaches it at 425M (42.5\%, labelled)}

Our compute budget allows replicating only one alternative, so we choose \textit{reverse periodic}: it has the lowest $\mathcal{L}_{\mathrm{LM}}$ and is the closest to \textit{standard periodic}. We rerun it and the baseline in two new conditions: Qwen3-4B on a newly sampled training set, and Granite-4.1-3B, which differs in architecture, tokenizer, and pretraining. Figure~\ref{fig:distillation-eval-curves}b shows the Granite runs; the second Qwen run and all training-loss curves are in Appendix~\ref{app:training_app}. 
The improvement over \textit{standard periodic} replicates in both conditions. Resampling the data moves the Qwen validation curves by only about a tenth of the gap between the two placements. At the end of training, neither placement has converged, and the gap between them shrinks by less than a tenth over the last quarter of training. Notably with Granite, \textit{reverse periodic} ends within 0.01 nats of the teacher's validation $\mathcal{L}_{\mathrm{LM}}$.

% \paragraph{Convergence.}
% : held-out KL and cross-entropy are still falling for every placement over the last quarter of training, so multilingual training on larger scale would likely improve all students further.
% But, as mentioned above, 1B tokens is plenty to see that the alternatives learn much faster. On validation set, their lead over \textit{standard periodic} appears within the first eval step and holds at every later step. As displayed in Figure~\ref{fig:distillation-eval-curves}, each alternative reaches \textit{standard periodic}'s final cross-entropy value in 40\% less tokens; see all training curves in Appendix~\ref{app:training_app}).
% We stop at 1B tokens, and no run has converged by then: held-out KL and cross-entropy are still falling for every placement over the last quarter of training, so multilingual training on larger scale would likely improve all students further.
% in case of extreme space hacking, this paragraph can be deleted, since this is extra rationale to back up our selection of reverse periodic as the candidate for further inspection apart from our heuristic intuition

% ; see all training curves in Appendix~\ref{app:training_app}.

% \paragraph{Better modeling without losing fidelity.}
% Every alternative placement models held-out text better than \textit{standard periodic} and stays closer to its teacher (Table~\ref{tab:distill-main}): for Qwen3-4B, both the cross-entropy gap and the relative KL drop by roughly a third.
% The lower cross-entropy therefore does not come from drifting away from the teacher.

\paragraph{Per-language gains.}
The $\mathcal{L}_{\mathrm{LM}}$ gains are achieved in almost every language, but to varying degrees. \textit{Reverse periodic} lowers perplexity relative to \textit{standard periodic} in 23 of the 26 languages with Qwen3-4B, in both runs, and in 25 with Granite; the perplexity of the few exceptions rises by less than 0.7\%. The largest gains, 3--13\%, come from Telugu, Malayalam, Tamil, Burmese, and Khmer, which rank among the top seven under both teachers, while the median language improves by about 1\% (Table~\ref{tab:result_perlang}). On $D_{\mathrm{KL}}$, \textit{reverse periodic} is closer to the teacher than \textit{standard periodic} in every language with Qwen3-4B and in all but two with Granite, and of the five placements only \textit{sandwich} is slightly closer on average. Here the pattern across languages is sharpest: under both teachers, the eight languages with the largest $D_{\mathrm{KL}}$ reduction are the same, namely Telugu, Malayalam, Tamil, Burmese, Khmer, Tibetan, Amharic, and Armenian, spanning all three sampling tiers and, importantly, all written in non-Latin scripts. 

% Most languages improve on both metrics, although by different amounts (Appendix Table~\ref{tab:result_perlang}). The improvement also survives two robustness checks. First, retraining \textit{standard periodic} and \textit{reverse periodic} on a different sample of the training data shifts their averages by only a small fraction of the gap between them. Second, with Granite-4.1-3B as the teacher, which differs in architecture, tokenizer, and pretraining data, \textit{reverse periodic} again beats \textit{standard periodic} on both metrics.

\paragraph{First-layer Full Attention}
The four alternatives have little in common except for all starting with a full attention layer.
Two keep the full-attention layers periodic or evenly spaced, and two cluster most of them at the ends of the network. Yet they end up close to one another and far from \textit{standard periodic} (Figure~\ref{fig:distillation_layouts}). This common property 
%they share which \textit{standard periodic} lacks, is a full-attention first layer.
is what \textit{standard periodic} lacks.
The fact that non-Latin scripts display the biggest differences suggests that the explanation may relate to detokenization, which is the first layer's most important role in multilingual NLU.

Crucially, this finding of first-layer importance is unlikely to come from the distillation setting itself. The layer selection in \citet{li2026distilling}, whose conversion is closest to ours (comparable initialization and forward $D_{\mathrm{KL}}$ training), \emph{never} selects the first layer as full attention \emph{on English data} (see Figure 6 in that paper).

\begin{table}[H]
\caption{Test results after 1B tokens, averaged over 26 languages. $R$: $D_{\mathrm{KL}}$ to the teacher relative to \textit{standard periodic} with the same teacher and training set (\eqref{eq:rel-kl}); $\bar\delta$: mean $\mathcal{L}_{\mathrm{LM}}$ gap to the teacher in nats (\eqref{eq:ce-gap}). For Qwen3-4B-Base, \textit{standard periodic} and \textit{reverse periodic} are trained twice on independently resampled data, shown as run 1 / run 2.}
\label{tab:distill-main}
\centering
\small
\setlength{\tabcolsep}{3pt}
\renewcommand{\arraystretch}{1.2}
\begin{tabular}{cccc}
\textbf{Teacher} & \textbf{Placement} & \textbf{relative $D_{\mathrm{KL}}$ $R$ $\downarrow$} & \textbf{$\mathcal{L}_{\mathrm{LM}}$ Gap $\bar\delta$ $\downarrow$} \\
\hline
Qwen3-4B-Base & \textit{Standard Periodic} & 1.000 / 1.000 & 0.071 / 0.068 \\
"" & \textit{Reverse Periodic} & 0.677 / 0.697 & \textbf{0.045} / 0.045 \\
"" & \textit{Sandwich} & \textbf{0.665} & 0.050 \\
"" & \textit{Interleaved Sandwich} & 0.689 & 0.049 \\
"" & \textit{Endpoint Spread} & 0.722 & 0.049 \\
\hline
Granite-4.1-3B-Base & \textit{Standard Periodic} & 1.000 & 0.042 \\
"" & \textit{Reverse Periodic} & \textbf{0.767} & \textbf{0.020} \\
\end{tabular}
\end{table}

\section{Conclusion and Open Questions}

This work provides the first analysis of the interaction between hybrid attention architectures and multilinguality.
We identify two concrete takeaways for model developers. First, multilingual LLMs should use hybrid attention because of the massive efficiency gains with minimal performance cost on long multilingual sequences. Second, our experimental setting suggests that LLMs learn multilingual data significantly better if the first decoder layer uses full attention. While our experiments display major training performance differences, it is, after all, distillation training on 1B tokens on small LLMs with one recurrent architecture. Regardless, our resource-limited conditions clearly prompts the theory that multilingual LLMs should start with full attention.

% Some limitation ideas from Lionel's notes:
% 1. distillation is framed under the scope of 1B total tokens. With extended training, there is a chance for double-descent.
% 2. Never tried any models that is bigger than 4B
% 3. Other dominant recurrent networks reported in RADLADS and other work (e.g. Mamba, lightning attention, etc.) is not tested in our experiment, worth documenting their dynamics
% 4. Due to constraint budget, we cannot really distill a model that we can confidently test against the downstream tasks, and thus not the cka (since also the training set is of different language coverage compared with the flores language set we picked)

Beyond this, Section~\ref{sec:cross-lingual-analysis} generally identifies that the hybridization of the attention blocks leads to some sort of representational reorganization. However, a mechanistic explanation of \emph{what} is actually occurring remains an open question. In general, mechanistic explanations remain an inhibitive challenge in LLM interpretability \citep{hendrycks2024misguided,sharkey2025open,nanda2025pragmatic}. For now, we can conclude that the inductive bias of each block matters and that the LLMs learn to reserve some sort of multilingual-relevant task for full attention layers.

\subsubsection*{Acknowledgments}

This research was made possible by financial support from the Amazon AI PhD Fellowship.

The authors acknowledge the excellent LLM architecture explanations at
the \href{https://sebastianraschka.com/llm-architecture-gallery/}{LLM Architecture Gallery}, curated by
Sebastian Raschka, as particularly helpful for this work. The authors also acknowledge Andrea Caciolai for discussions on modern LLM architectures.

\hyperref[models_details]{Click here to access the Appendix.}

% \begingroup
% \raggedright
% \raggedbottom
\bibliography{custom}
\bibliographystyle{iclr2027_conference}
\clearpage
% \endgroup

\appendix\counterwithin{figure}{section}\counterwithin{table}{section}

\section{Further Details about Models} \label{models_details}

\begin{table}[h]
\caption{Continuing on from Table~\ref{tab:model-comparison}, we provide additional architecture details. ``Position'' is the positional embeddings used \emph{in the full attention layers}.}
\label{tab:model-details}
\centering
\small
\setlength{\tabcolsep}{3pt}
\renewcommand{\arraystretch}{1.2}
\begin{tabular}{ccccccc}
\textbf{Model Name} & \textbf{Citation} & \textbf{\# Layers} & \textbf{MoE ?} & \textbf{Params} & \textbf{Hidden Size} & \textbf{Position} \\
\hline
Qwen3.5-35B-A3B & \citet{qwen35} & 40 & Yes & 35B & 2048 & RoPE \\
Qwen3-30B-A3B & \citet{yang2025qwen3technicalreport} & 48 & "" & 30B & "" & "" \\
\hline
\olmo-Hybrid & \citet{merrill2026olmo} & 32 & No & 7B & 3840 & RoPE \\
\olmo-3-7B & \citet{olmo2026olmo3} & "" & "" & "" & 4096 & RoPE \& YaRN\\
\hline
Ring-mini-linear-2.0 & \citet{ring} & 20 & Yes & 16B & 2048 & RoPE \\ 
Ring-mini-2.0 & \citet{lingteam2025stepevolvesscalingreinforcement} & "" & "" & "" & "" & "" \\
\hline
Granite-4.0-H-Micro & \citet{granite2025} & 40 & No & 3B & 2048 & None (NoPE) \\
Granite-4.0-Micro & "" & "" & "" & "" & 2560 & RoPE \\
\hline
Ling-2.6-flash & \cite{ling26} & 32 & Yes & 107B & 4096 & Partial RoPE \\
\hline
Tiny Aya Global & \citet{tinyaya} & 36 & No & 3B & 2048 & RoPE \\
\end{tabular}
\end{table}

\begin{table}[h]
\caption{Architectural Components}
\centering
\small
\setlength{\tabcolsep}{3pt}
\renewcommand{\arraystretch}{1.2}
\begin{tabular}{ccc}
\textbf{Display Name} & \textbf{Full Name} & \textbf{Citation}  \\
\hline
GQA & Grouped Query Attention & \citet{ainslie-etal-2023-gqa} \\
Gated GQA & --- & \citet{qiu2025gated} \\
MLA & Multi-Head Latent Attention & \citet{deepseekv2} \\
SWA & Sliding Window Attention & \citet{longformer} \\
\hline
Gated DeltaNet & --- & \citet{gdn} \\
Gated DeltaNet w/ Negative Eigenvalues & --- & \citet{grazzi2025unlocking} \\
Lightning Attn 2 & --- & \citep{qin2024lightningattention2freelunch} \\
Mamba-2 & --- & \citet{mamba2} \\
\hline
RoPE & Rotary Position Embedding & \citep{rope} \\
YaRN & Yet another RoPE extensioN & \citep{yarn} \\
NoPE & No positional embedding & \citep{nope} \\
\end{tabular}
\end{table}

While we do not focus on it in this work, the models all use various formulas for positional embeddings across attention blocks. This likely has some degree of consequence on the progression of multilingual representations in heterogeneous LLMs.
\section{Multilingual Task Evaluations} \label{benchmarks}

The first step of our analysis was to evaluate the pairs of hybrid/non-hybrid models on multilingual benchmarks. We evaluate the four model pairs on MGSM \citep{shi2023language}, MMLU ProX \citep{xuan-etal-2025-mmlu-prox}, Belebele \citep{bandarkar-etal-2024-belebele}, and Global-PIQA \citep{chang2025globalpiqa}. However, as discussed in Section~\ref{performance}, comparisons are severely limited by the lack of controlled comparability and transparency into model development. We therefore evaluate performance relative to English as a measure of cross-lingual transfer. The results are noisy and inconsistent across models, so we do not report them in detail because the investigation was inconclusive.
\section{Design of Distillation Experiment} \label{distillation}

\subsection{Data}
\label{app:data}
\begin{table}[h]
\caption{Distillation mixture ($N = 10^9$ tokens). Shares are token quotas under each teacher's tokenizer. English is drawn from FineWeb (sample-100BT) and all other languages from FineWeb-2; codes follow FineWeb-2 (ISO 639-3 and script). }
\label{tab:distill-data}
\centering
\small
\setlength{\tabcolsep}{3pt}
\renewcommand{\arraystretch}{1.2}
\begin{tabular}{cccc}
\textbf{Tier} & \textbf{Languages} & \textbf{Share Each} & \textbf{Tokens Each} \\
\hline
English & eng\_Latn & 30.0\% & 300M \\
\hline
\textit{High} (8) & rus\_Cyrl, hin\_Deva, cmn\_Hani, arb\_Arab & 4.08\% & 40.8M \\
 & ind\_Latn, vie\_Latn, tur\_Latn, tam\_Taml &  &  \\
\hline
\textit{Mid} (14) & ell\_Grek, hye\_Armn, yue\_Hani, mya\_Mymr, heb\_Hebr & 2.33\% & 23.3M \\
 & amh\_Ethi, hau\_Latn, zsm\_Latn, fil\_Latn, tel\_Telu &  &  \\
 & mal\_Mlym, azj\_Latn, kaz\_Cyrl, khm\_Khmr &  &  \\
\hline
\textit{Low} (4) & gle\_Latn, bod\_Tibt, uig\_Arab, mnw\_Mymr$^\ast$ & 1.17\% & 11.7M \\
\end{tabular}
\end{table}

\paragraph{Data sampling recipe.}
Training data span 27 languages from seven families (Indo-European, Sino-Tibetan, Afro-Asiatic, Austronesian, Dravidian, Turkic, and Austro-Asiatic). Within each family, we choose languages that differ in script (Table~\ref{tab:distill-langs}). English receives a fixed 30\% of the 1B-token budget as an anchor; the remaining 70\% is split across three sampling tiers based on language resource levels with per-language shares in a 3.5\,:\,2\,:\,1 ratio.

\begingroup
\setlength{\tabcolsep}{3pt}
\renewcommand{\arraystretch}{1.2}
\setlength{\LTcapwidth}{\textwidth}
\begin{longtable}{cccccc}
\caption{Distillation languages, grouped by family. \emph{FW2 Docs} is the number of FineWeb-2 training documents \citep{penedo2025fineweb2pipelinescale}; English is drawn from FineWeb \citep{penedo2024finewebdatasetsdecantingweb}. \emph{Tier} sets each language's share of the $10^9$-token budget: English 30\%, High 4.08\%, Mid 2.33\%, Low 1.17\% (a 3.5\,:\,2\,:\,1 ratio), i.e., 300M, 40.8M, 23.3M, and 11.7M tokens under each teacher's tokenizer.}
\label{tab:distill-langs}\\
\textbf{Family} & \textbf{Language} & \textbf{Code} & \textbf{Script Type} & \textbf{Tier} & \textbf{FW2 Doc \#} \\
\hline
\endfirsthead
\caption[]{(continued)}\\
\textbf{Family} & \textbf{Language} & \textbf{Code} & \textbf{Script Type} & \textbf{Tier} & \textbf{FW2 Docs} \\
\hline
\endhead
Indo-European & English & eng\_Latn & Alphabet & Anchor & --- \\
 & Russian & rus\_Cyrl & Alphabet & High & 699.1M \\
 & Hindi & hin\_Deva & Abugida & High & 22.1M \\
 & Greek & ell\_Grek & Alphabet & Mid & 47.4M \\
 & Armenian & hye\_Armn & Alphabet & Mid & 1.8M \\
 & Irish & gle\_Latn & Alphabet & Low & 0.65M \\
\hline
Sino-Tibetan & Mandarin & cmn\_Hani & Logographic & High & 636.1M \\
 & Cantonese & yue\_Hani & Logographic & Mid & 0.31M \\
 & Burmese & mya\_Mymr & Abugida & Mid & 1.6M \\
 & Tibetan & bod\_Tibt & Abugida & Low & 0.16M \\
\hline
Afro-Asiatic & Arabic & arb\_Arab & Abjad & High & 62.0M \\
 & Hebrew & heb\_Hebr & Abjad & Mid & 14.5M \\
 & Amharic & amh\_Ethi & Abugida & Mid & 0.43M \\
 & Hausa & hau\_Latn & Alphabet & Mid & 0.57M \\
\hline
Austronesian & Indonesian & ind\_Latn & Alphabet & High & 100.2M \\
 & Malay & zsm\_Latn & Alphabet & Mid & 9.4M \\
 & Filipino & fil\_Latn & Alphabet & Mid & 2.3M \\
\hline
Dravidian & Tamil & tam\_Taml & Abugida & High & 5.5M \\
 & Telugu & tel\_Telu & Abugida & Mid & 2.0M \\
 & Malayalam & mal\_Mlym & Abugida & Mid & 3.3M \\
\hline
Turkic & Turkish & tur\_Latn & Alphabet & High & 95.1M \\
 & Azerbaijani & azj\_Latn & Alphabet & Mid & 7.3M \\
 & Kazakh & kaz\_Cyrl & Alphabet & Mid & 3.3M \\
 & Uyghur & uig\_Arab & Alphabet & Low & 0.17M \\
\hline
Austro-Asiatic & Vietnamese & vie\_Latn & Alphabet & High & 61.1M \\
 & Khmer & khm\_Khmr & Abugida & Mid & 1.6M \\
 & Mon & mnw\_Mymr & Abugida & Low & 2.3K \\
\end{longtable}
\endgroup

\paragraph{Held-out sets and replicate runs.}
For each language, we fix a canonical document order by shuffling its FineWeb-2 stream (shard order and a 2,000-document buffer) with a seed held constant across all runs. The first 2,000 documents in this order are never trained on: documents 1--1,000 form the validation set, evaluated every $10^7$ training tokens, and documents 1,001--2,000 form the test set, evaluated once after training. Because this reservation does not depend on a run's data seed, all runs with the same teacher are scored on identical validation and test token sequences. Training data are drawn from the remaining documents: the data seed reshuffles each language's remaining stream and the order in which languages are interleaved in the training set, so different seeds train on largely different documents. The only exception is Mon, whose 340 remaining documents are reused unchanged. 

To distinguish data seeding with the initialization seed reported in Table~\ref{tab:distill-hparams}, the new recurrent parameters are always initialized with seed 0. We train \textit{standard periodic} and \textit{reverse periodic} a second time with data seed 1, so the two runs differ only in the training sample and estimate variability due to data rather than initialization. 

\subsection{Hyperparameter}
\label{app:hyperparameter}
\newcommand{\both}[1]{\multicolumn{2}{p{0.63\linewidth}@{}}{#1}}
\begingroup
\setlength{\tabcolsep}{3pt}
\renewcommand{\arraystretch}{1.2}
\setlength{\LTcapwidth}{\textwidth}
\begin{longtable}{ccc}
\caption{Distillation hyperparameters and experimental details. Symbols follow Algorithm~\ref{alg:hybrid-distill} and Eq.~\eqref{eq:lr}.}
\label{tab:distill-hparams}\\
\textbf{Hyperparameter} & \textbf{Qwen3-4B-Base} & \textbf{Granite-4.1-3B-Base} \\
\hline
\endfirsthead
\caption[]{(continued)}\\
\textbf{Hyperparameter} & \textbf{Qwen3-4B-Base} & \textbf{Granite-4.1-3B-Base} \\
\hline
\endhead
 & \citet{yang2025qwen3technicalreport} & \citet{granite2026} \\
\# Layers (full / recurrent) & 36 (9 / 27) & 40 (10 / 30) \\
Hidden size $d$ & 2560 & 2560 \\
Heads $H$ / $H_{kv}$ & 32 / 8 & 40 / 8 \\
Head dim $d_h$ & 128 & 64 \\
Recurrent state per layer ($H d_h^2$) & 524K & 164K \\
Vocabulary size & 151,936 & 100,352 \\
Orderings & All five & \textit{standard periodic}, \textit{reverse periodic} \\
\hline
Recurrent attention & \multicolumn{2}{c}{Gated DeltaNet, conv.\ width 4 (SiLU), output gate} \\
Objective & \multicolumn{2}{c}{Forward $D_{KL}$, $\tau = 1$} \\
Layer selection & \multicolumn{2}{c}{prescribed $\mathcal{F}$} \\
LM head & \multicolumn{2}{c}{Untied (separate trainable copy of $E$)} \\
\hline
Tokens $N$ / steps $S$ & \multicolumn{2}{c}{1B (single pass) / 10,172} \\
Batch $B \times n$ & \multicolumn{2}{c}{$96 \times 1024$ (micro-batch 2, accumulation 48)} \\
Optimizer & \multicolumn{2}{c}{AdamW, 8-bit paged states} \\
$(\beta_1, \beta_2)$ / $\epsilon_{\mathrm{Adam}}$ / weight decay & \multicolumn{2}{c}{$(0.9, 0.999)$ / $10^{-8}$ / 0.01} \\
Peak LR $\hat\eta_{\mathrm{rec}}$ / $\hat\eta_{\mathrm{rest}}$ & \multicolumn{2}{c}{$7 \times 10^{-5}$ / $2 \times 10^{-5}$} \\
Warmup $s_w$ (start factor $\epsilon$) & \multicolumn{2}{c}{305 steps, 3\% ($10^{-3}$)} \\
Decay & \multicolumn{2}{c}{Cosine to $\eta_{\min} = 10^{-5}$} \\
Gradient norm clipping $c$ & \multicolumn{2}{c}{1.0} \\
Precision & \multicolumn{2}{c}{bf16, gradient checkpointing} \\
Init.\ seed (GDN parameters) & \multicolumn{2}{c}{0} \\
\hline
Languages & \multicolumn{2}{c}{27 (Table~\ref{tab:distill-data})} \\
Held-out docs per language & \multicolumn{2}{c}{1000 validation + 1000 test} \\
Held-out tokens per language & \multicolumn{2}{c}{validation: $16 \times 1024$, test: $32 \times 1024$} \\
Validation frequency & \multicolumn{2}{c}{Every 10M training tokens} \\
Metric & \multicolumn{2}{c}{Per-language forward $D_{KL}$ (nats/token)} \\
\hline
Software & \multicolumn{2}{c}{PyTorch \;2.10.0 (CUDA 12.6)\; , \;Transformers \;4.56.2} \\*
& \multicolumn{2}{c}{FLA \;0.5.2\; , \;bitsandbytes \;0.50.0} \\
\end{longtable}
\endgroup

\paragraph{Weight Tying.}
Qwen3-4B-Base and Granite-4.1-3B-Base both tie the input embedding and the LM head. Following a workaround that \citet{goldstein2026radladsrapidattentiondistillation} suggest for smaller tied models, we untie them during distillation, training the head as a separate copy of the embedding to improve learning; We evaluate this untied student directly, since it is the model the distillation objective optimizes; unlike RADLADS, we do not re-tie the head afterwards.

\subsection{Algorithm of distillation}

\paragraph{Gated DeltaNet parameterization}
In each recurrent attention layer, head $h$ keeps a state $\mathbf{M}_t \in \mathbb{R}^{d_h \times d_h}$ updated by the gated delta rule \citep{gdn}:
\begin{equation}
\mathbf{M}_t = \alpha_t\, \mathbf{M}_{t-1}\big(I - \beta_t\, k_t k_t^{\top}\big) + \beta_t\, v_t k_t^{\top}, \qquad o_t = \mathbf{M}_t\, q_t,
\end{equation}
where $q_t, k_t, v_t$ are the projections $W_Q x_t, W_K x_t, W_V x_t$ of RMSNormed input $x_t$ passed through a causal depthwise convolution of width $K$ and SiLU, with $q_t$ and $k_t$ $\ell_2$-normalized. The decay and write strength are scalars per head that depend on the input: $\alpha_t = \exp\!\big(-e^{A_{\log}}\,\mathrm{softplus}(W_a x_t + b_\Delta)\big)$ and $\beta_t = \sigma(W_\beta x_t)$. Head outputs pass through an RMSNorm with scale $\gamma$, gated by $\mathrm{SiLU}(W_z x_t)$, before $W_O$.

\begin{algorithm}[H]
\caption{Hybrid distillation with a prescribed full attention placement}
\label{alg:hybrid-distill}
\begin{algorithmic}[1]
\Require teacher $\mathcal{M}_{\mathrm{T}}$: $L$ layers, width $d$, $H$ query / $H_{kv}$ KV heads of dim.\ $d_h$
\Require tied embedding and LM head $E \in \mathbb{R}^{|\mathcal{V}| \times d}$
\Require softmax-attention set $\mathcal{F} \subset \{0,\dots,L-1\}$, $|\mathcal{F}| = L/4$ \Comment{the ordering}
\Require mixture $\mathcal{D}$, budget $N$, batch $B \times n$; $\tau$, $c$, $\hat\eta_{\mathrm{rec}}$, $\hat\eta_{\mathrm{rest}}$, $\eta_{\min}$ (Table~\ref{tab:distill-hparams})
\Ensure hybrid student $\mathcal{M}_{\mathrm{S}}$
\Statex \textbf{Initialization}
\State $\mathcal{M}_{\mathrm{S}} \gets \mathcal{M}_{\mathrm{T}}$; freeze $\mathcal{M}_{\mathrm{T}}$
\For{$\ell \notin \mathcal{F}$} \Comment{layers that become recurrent}
    \State $\mathrm{Attn}_\ell \gets \mathrm{GDN}_\ell$ with $H$ heads of dimension $d_h$
    \State $W_Q, W_O \gets W_Q^{\mathrm{T}}, W_O^{\mathrm{T}}$
    \State $W_K^{(h)} \gets W_K^{\mathrm{T},(\lfloor h/G \rfloor)}$, $W_V^{(h)} \gets W_V^{\mathrm{T},(\lfloor h/G \rfloor)}$, $h = 0,\dots,H-1$ \Comment{$G = H/H_{kv}$}
    \State $\kappa_j \gets \delta_{j,K}$ in every channel \Comment{identity short conv, width $K$}
    \State $A_{\log} \gets -4 \cdot \mathbf{1}_H$ \Comment{decay $\alpha_t \approx 1$}
    \State $W_a, W_\beta, W_z \sim \mathcal{U}\big(-d^{-1/2}, d^{-1/2}\big)$; \ $\gamma \gets \mathbf{1}$
    \State $b_\Delta \gets \mathrm{softplus}^{-1}(\Delta)$, \ $\log \Delta_h \overset{\text{iid}}{\sim} \mathcal{U}(\log 10^{-3}, \log 10^{-1})$
\EndFor
\State $W_{\mathrm{head}} \gets \mathrm{copy}(E)$ \Comment{untie: separate trainable copy}
\State $\theta_{\mathrm{rec}} \gets \bigcup_{\ell \notin \mathcal{F}} \theta(\mathrm{GDN}_\ell)$; \ $\theta_{\mathrm{rest}} \gets \theta(\mathcal{M}_{\mathrm{S}}) \setminus \theta_{\mathrm{rec}}$
\State $S \gets \lfloor N / (Bn) \rfloor$
\Statex \textbf{Distillation}
\For{$s = 1, \dots, S$}
    \State $X \sim \mathcal{D}$, \ $X \in \mathcal{V}^{B \times n}$ \Comment{packed sequences}
    \State $p_{\mathrm{T}} \gets \mathrm{softmax}\big(\mathcal{M}_{\mathrm{T}}(X)/\tau\big)$, \ $p_{\mathrm{S}} \gets \mathrm{softmax}\big(\mathcal{M}_{\mathrm{S}}(X)/\tau\big)$
    \State $\mathcal{L} \gets \frac{\tau^2}{Bn} \sum_{b,i} D_{KL}\big(p_{\mathrm{T}}[b,i] \,\|\, p_{\mathrm{S}}[b,i]\big)$
    \State $\mathbf{g} \gets \nabla_{\theta} \mathcal{L}$; \ $\mathbf{g} \gets \mathbf{g} \cdot \min\big(1, c / \|\mathbf{g}\|_2\big)$
    \State $\theta_k \gets \mathrm{AdamW}\big(\theta_k, \mathbf{g}_k; \eta_k(s)\big)$, \ $k \in \{\mathrm{rec}, \mathrm{rest}\}$ \Comment{Eq.~\eqref{eq:lr}}
\EndFor
\State \Return $\mathcal{M}_{\mathrm{S}}$
\end{algorithmic}
\end{algorithm}

\paragraph{Learning rate configuration.} We train with two parameter groups, following RADLADS \citep{goldstein2026radladsrapidattentiondistillation}: the converted recurrent attentions, with learning rate $\eta_{\mathrm{rec}}$, and all remaining parameters (MLPs, embeddings, normalization layers, and the retained full-attention layers), with a lower rate $\eta_{\mathrm{rest}}$. The lower rate keeps the MLPs and embeddings, where most of the teacher's knowledge resides, from drifting during conversion. 

The schedule follows HALO's distillation stage \citep{chen2026hybridlinearattentionright}: a cosine decay to $10^{-5}$ over 1B tokens, with a peak rate chosen by model size ($10^{-4}$ at 2B and $5 \times 10^{-5}$ at 5B in HALO). For both of our teachers we use peak rates $\eta_{\mathrm{rec}} = 7 \times 10^{-5}$ and $\eta_{\mathrm{rest}} = 2 \times 10^{-5}$, reached after a linear warmup over the first 3\% of steps; both groups decay along the same cosine schedule.

\begin{equation}
\eta_k(s) =
\begin{cases}
\hat\eta_k \left(\epsilon + (1-\epsilon)\,\dfrac{s}{s_w}\right), & s \le s_w, \\[6pt]
\eta_{\min} + \dfrac{\hat\eta_k - \eta_{\min}}{2}\left(1 + \cos\dfrac{\pi (s - s_w)}{S - s_w}\right), & s > s_w,
\end{cases}
\qquad k \in \{\mathrm{rec}, \mathrm{rest}\}
\label{eq:lr}
\end{equation}

\subsection{Training }
\label{app:training_app}
\paragraph{Training protocol.}
Every run follows the same recipe; only the full attention set $\mathcal{F}$ changes between orderings (Algorithm~\ref{alg:hybrid-distill}). Starting from the teacher, we convert the layers outside $\mathcal{F}$ to Gated DeltaNet and train the whole student in a single stage using forward $D_{KL}$ to the frozen teacher, omitting the hidden-state alignment stage and data-driven layer selection of prior conversion pipelines \citep{goldstein2026radladsrapidattentiondistillation, chen2026hybridlinearattentionright, li2026distilling}. Each run makes one pass over $10^9$ tokens of the mixture in Table~\ref{tab:distill-langs}, packed into sequences of 1024 tokens, with the hyperparameters in Table~\ref{tab:distill-hparams}. The new recurrent parameters are always initialized with seed 0; the replicate runs differ only in the training sample (Appendix~\ref{app:data}). Each run uses about 99 hours with Qwen3-4B-Base or 82 hours with Granite-4.1-3B-Base on a single NVIDIA L40S (48GB).

\begin{figure}[H]
  \centering
  \includegraphics[width=0.75\linewidth]{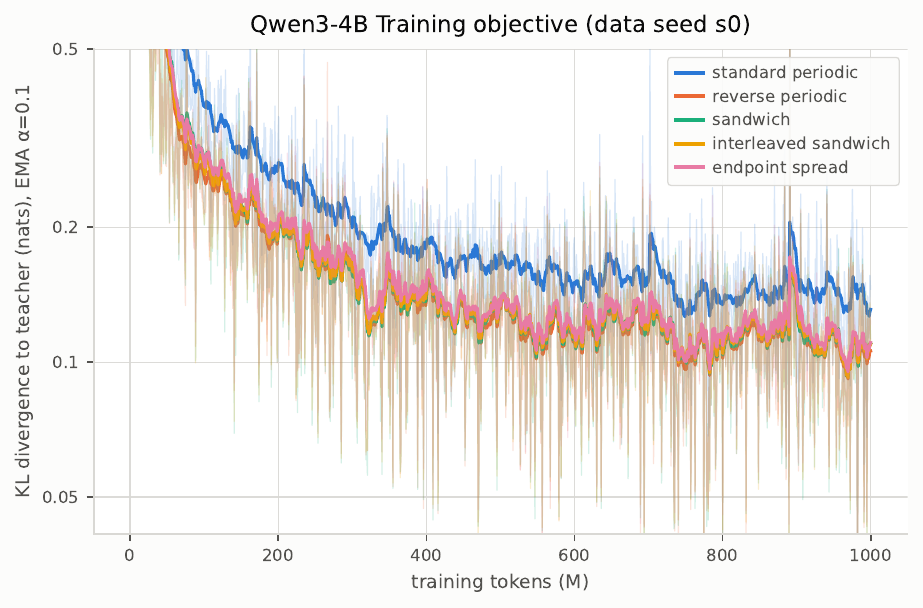}
  \caption{Distillation training curves for Qwen3-4B with data seed 0, one run per placement. Faint lines show the per-step $D_{KL}$ on the training batch and bold lines its exponential moving average ($\alpha = 0.1$). The y-axis is logarithmic and cut off at 0.5 nats; every run starts between 7.0 and 10.5 nats.}
  \label{fig:train-qwen-s0}
\end{figure}

\begin{figure}[H]
    \centering
    \begin{minipage}[t]{0.49\linewidth}
        \centering
        \includegraphics[width=\linewidth]{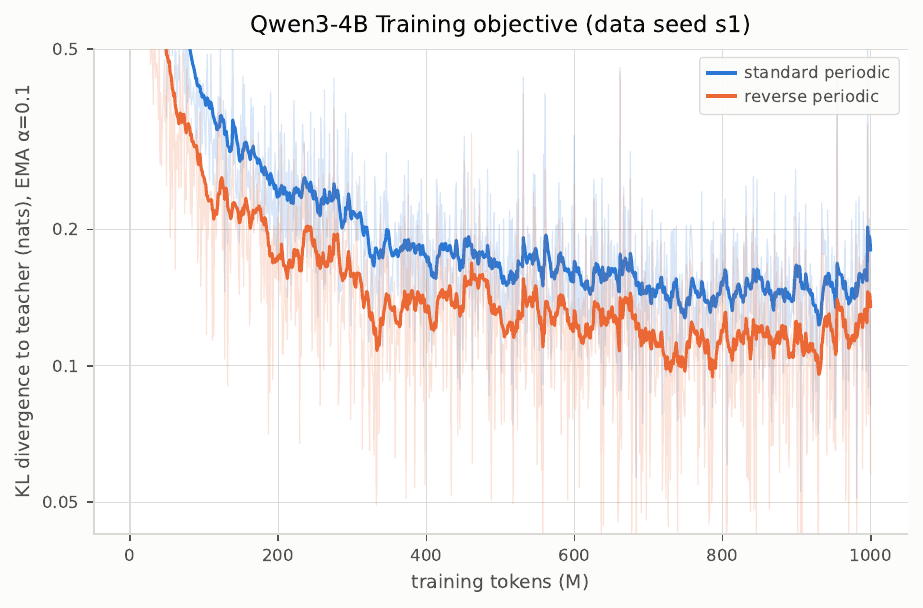}
        \small (a) Qwen3-4B student model
    \end{minipage}
    \hfill
    \begin{minipage}[t]{0.49\linewidth}
        \centering
        \includegraphics[width=\linewidth]{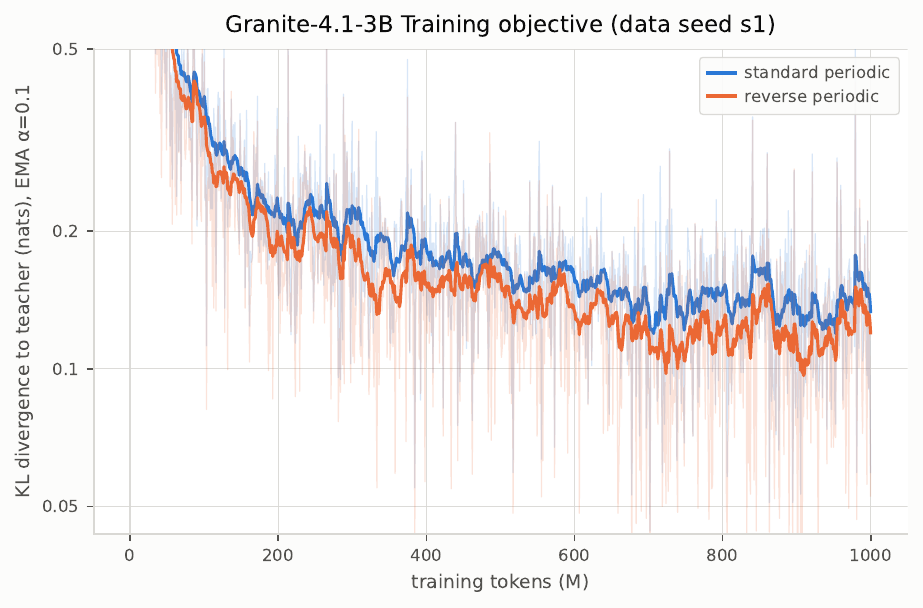}
        \small (b) Granite-4.1-3B student model
    \end{minipage}
    \caption{Distillation training curves for the Qwen3-4B and Granite-4.1-3B \textit{standard periodic} and \textit{reverse periodic} placement with data seed 1.}
    \label{fig:distillation-training-curves}
\end{figure}

\paragraph{Gated DeltaNet initialization.}
We initialize each converted layer by transferring the teacher's attention weights \citep{goldstein2026radladsrapidattentiondistillation, chen2026hybridlinearattentionright}. The Gated DeltaNet layer adopts the teacher's head geometry ($H$ heads of dimension $d_h$, with value dimension $d_h$), so $W_Q$ and $W_O$ are copied directly. Both teacher models use grouped-query attention, whereas Gated DeltaNet has one key and value head pair per query head, so we clone each key and value head across its query group, $W_K^{(h)} \gets W_K^{\mathrm{T},(\lfloor h/G \rfloor)}$ and likewise for $W_V$, as in HALO. Parameters without a teacher counterpart are set so that each converted layer starts with no convolutional mixing and almost no forgetting. The remaining parameters ($W_a$, $W_\beta$, $W_z$, $b_\Delta$, $\gamma$) use the default initialization of the FLA implementation (Algorithm~\ref{alg:hybrid-distill}). Retained attention layers, MLPs, norms, and embeddings are copied from the teacher unchanged.

\subsection{Metrics and Results}

\paragraph{Evaluation protocol.}
All metrics are computed on held-out documents. The validation split is evaluated every $10^7$ training tokens and produces all training curves; the test split is evaluated once, after training, and produces all final numbers. Each language contributes $16 \times 1024$ validation and $32 \times 1024$ test tokens, packed into full sequences of the training length $n = 1024$ and no longer contexts are evaluated. $D_{\mathrm{KL}}$ and cross-entropy $\mathcal{L}_{\mathrm{LM}}$ are averaged over all tokens of a language; the aggregate metrics then average over the 26 languages in $\Lambda$, which include English and exclude Mon\footnote{Mon has only about 2.3K documents in FineWeb-2, so reserving 2{,}000 for validation and test leaves roughly 340 training documents which are repeated $\approx5$ times to fill its sampling quota; the reshuffling on each repetition could leak reserved documents into training. We therefore keep Mon in the training mixture but exclude it from all validation and test averages.} (Table~\ref{tab:distill-langs}). Test split is evaluated once after 1B distillation tokens. All results are for the untied student (Appendix~\ref{app:hyperparameter}).

\paragraph{Cross-entropy gap to the teacher.}
We report the mean per-token $\mathcal{L}_{\mathrm{LM}}$ gap between student and teacher on held-out text, averaged over the evaluation languages $\lambda \in \Lambda$ ($|\Lambda| = 26$):
\begin{equation}
\bar\delta_{\mathrm{S}}(s) = \frac{1}{|\Lambda|}\sum_{\lambda \in \Lambda}\big(\mathcal{L}_{\mathrm{LM},\lambda}^{\mathrm{S}}(s) - \mathcal{L}_{\mathrm{LM},\lambda}^{\mathrm{T}}\big),
\label{eq:ce-gap}
\end{equation}
where $\mathcal{L}_{\mathrm{LM},\lambda}^{\mathrm{S}}(s)$ is the student's per-token $\mathcal{L}_{LM}$ (in nats) on language $\lambda$ after $s$ optimization steps and $\mathcal{L}_{\mathrm{LM},\lambda}^{\mathrm{T}}$ the teacher's; lower is better. Every language contributes the same number of held-out tokens, so $\bar\delta_{\mathrm{S}}$ also equals the per-token gap over the pooled evaluation set. By definition $\log \mathrm{ppl} = \mathcal{L}_{\mathrm{LM}}$, the gap has a direct perplexity reading:
\begin{equation*}
e^{\bar\delta_{\mathrm{S}}(s)} = \Big(\prod_{\lambda \in \Lambda} \frac{\mathrm{ppl}^{\mathrm{S}}_\lambda(s)}{\mathrm{ppl}^{\mathrm{T}}_\lambda}\Big)^{1/|\Lambda|},
\end{equation*}
the geometric mean of the student-to-teacher perplexity ratios, i.e. $\bar\delta_{\mathrm{S}} = 0.05$ means the student's perplexity is on average $\sim5\%$ above the teacher's.

\paragraph{$D_{\mathrm{KL}}$ relative to the baseline placement.}
To compare different placements, we measure each student's $D_{\mathrm{KL}}$ to the teacher relative to the \textit{standard periodic} student distilled from the same teacher, taking the geometric mean of the per-language ratios at the same optimization step $s$:
\begin{equation}
R_{\mathrm{S}}(s) = \exp\!\Big(\frac{1}{|\Lambda|}\sum_{\lambda \in \Lambda} \log \frac{D_{\mathrm{KL},\lambda}^{\mathrm{S}}(s)}{D_{\mathrm{KL},\lambda}^{\mathrm{B}}(s)}\Big),
\label{eq:rel-kl}
\end{equation}
where $D_{\mathrm{KL},\lambda}^{\mathrm{S}}(s)$ is the mean per-token $D_{\mathrm{KL}}(p_{\mathrm{T}} \,\|\, p_{\mathrm{S}})$ of student $\mathcal{M}_{\mathrm{S}}$ on the held-out text of language $\lambda$, and $D_{\mathrm{KL},\lambda}^{\mathrm{B}}(s)$ is the same quantity for the baseline. $R_{\mathrm{S}} = 1$ matches the baseline, and $R_{\mathrm{S}} < 1$ means the student is closer to the teacher than the baseline is (lower is better); i.e. $R_{\mathrm{S}} = 0.7$ corresponds to a 30\% lower $D_{\mathrm{KL}}$ in the geometric mean over languages.

\paragraph{Unit in nats per token.} All $\mathcal{L}_{\mathrm{LM}}$ and $D_{\mathrm{KL}}$ quantities are reported in nats per
token (natural logarithm), so that $\mathrm{ppl} = \exp(\mathcal{L}_{\mathrm{LM}})$. We use
nats mainly because bits in language-model evaluation conventionally
appear as \emph{bits per byte}, a tokenizer-invariant measure. Our
gap is per \emph{token} and is not tokenizer-invariant: Qwen3-4B and
Granite-4.1-3B segment the same text differently, so gaps are comparable within a
teacher's tokenizer but never across one.

\clearpage
\paragraph{$\mathcal{L}_{\mathrm{LM}}$ gap by language.}

\begingroup
\footnotesize
\setlength{\tabcolsep}{3pt}
\renewcommand{\arraystretch}{1.2}
\setlength{\LTcapwidth}{\textwidth}
\begin{longtable}{@{}lcc|rrrrr|rr|rr@{}}
\caption{Per-language test results. For each language, the $D_{\mathrm{KL}}$ row gives $D_{\mathrm{KL}}(p_{\mathrm{T}} \,\|\, p_{\mathrm{S}})$ (nats per token) and the $\mathcal{L}_{\mathrm{LM}}$ row the $\mathcal{L}_{\mathrm{LM}}$ gap to the teacher, $\mathcal{L}_{\mathrm{LM},\lambda}^{\mathrm{S}} - \mathcal{L}_{\mathrm{LM},\lambda}^{\mathrm{T}}$ (nats); lower is better for both, and a negative $\mathcal{L}_{\mathrm{LM}}$ gap means the student is less perplexed than the teacher. The last two rows give $R$ (\eqref{eq:rel-kl}), relative to SP of the run with the same data seed, and $\bar\delta$ (\eqref{eq:ce-gap}). Teachers are the base models (Qwen3-4B-Base, Granite-4.1-3B-Base). Languages are grouped by family as in Table~\ref{tab:distill-langs}. SP: \textit{Standard Periodic}; RP: \textit{Reverse Periodic}; SW: \textit{Sandwich}; IS: \textit{Interleaved Sandwich}; ES: \textit{Endpoint Spread}. Tier: E/H/M/L = English/High/Mid/Low sampling tier (Table~\ref{tab:distill-langs}). $^{\dagger}$Mon is excluded from all averages.}
\label{tab:result_perlang}\\
\toprule
\multicolumn{3}{c}{} & \multicolumn{7}{c}{\textbf{Qwen3-4B}} & \multicolumn{2}{c}{\textbf{Granite-4.1-3B}} \\
\multicolumn{3}{c}{} & \multicolumn{5}{c}{data seed 0} & \multicolumn{2}{c}{data seed 1} & \multicolumn{2}{c}{data seed 1} \\
\textbf{Language} & \textbf{Tier} & \textbf{Metric} & \textbf{SP} & \textbf{RP} & \textbf{SW} & \textbf{IS} & \textbf{ES} & \textbf{SP} & \textbf{RP} & \textbf{SP} & \textbf{RP} \\
\midrule
\endfirsthead
\caption[]{(continued)}\\
\toprule
\multicolumn{3}{c}{} & \multicolumn{7}{c}{\textbf{Qwen3-4B}} & \multicolumn{2}{c}{\textbf{Granite-4.1-3B}} \\
\multicolumn{3}{c}{} & \multicolumn{5}{c}{data seed 0} & \multicolumn{2}{c}{data seed 1} & \multicolumn{2}{c}{data seed 1} \\
\textbf{Language} & \textbf{Tier} & \textbf{Metric} & \textbf{SP} & \textbf{RP} & \textbf{SW} & \textbf{IS} & \textbf{ES} & \textbf{SP} & \textbf{RP} & \textbf{SP} & \textbf{RP} \\
\midrule
\endhead
English & E & $D_{\mathrm{KL}}$ & 0.160 & \textbf{0.140} & 0.148 & 0.147 & 0.144 & 0.160 & \textbf{0.142} & 0.212 & \textbf{0.203} \\*
 & & $\mathcal{L}_{\mathrm{LM}}$ & 0.119 & \textbf{0.106} & 0.116 & 0.112 & 0.108 & 0.120 & \textbf{0.108} & 0.157 & \textbf{0.151} \\
Russian & H & $D_{\mathrm{KL}}$ & 0.113 & \textbf{0.092} & 0.100 & 0.099 & 0.096 & 0.111 & \textbf{0.093} & 0.099 & \textbf{0.097} \\*
 & & $\mathcal{L}_{\mathrm{LM}}$ & 0.088 & \textbf{0.073} & 0.082 & 0.079 & 0.076 & 0.089 & \textbf{0.074} & 0.054 & \textbf{0.053} \\
Hindi & H & $D_{\mathrm{KL}}$ & 0.073 & \textbf{0.056} & \textbf{0.056} & 0.057 & 0.058 & 0.072 & \textbf{0.055} & 0.068 & \textbf{0.064} \\*
 & & $\mathcal{L}_{\mathrm{LM}}$ & 0.043 & 0.033 & 0.035 & 0.034 & \textbf{0.032} & 0.043 & \textbf{0.032} & 0.025 & \textbf{0.021} \\
Greek & M & $D_{\mathrm{KL}}$ & 0.084 & 0.066 & \textbf{0.062} & 0.065 & 0.071 & 0.085 & \textbf{0.067} & 0.073 & \textbf{0.067} \\*
 & & $\mathcal{L}_{\mathrm{LM}}$ & 0.050 & 0.034 & 0.033 & \textbf{0.032} & 0.039 & 0.050 & \textbf{0.037} & 0.003 & \textbf{0.001} \\
Armenian & M & $D_{\mathrm{KL}}$ & 0.090 & 0.061 & \textbf{0.055} & 0.059 & 0.066 & 0.090 & \textbf{0.061} & 0.122 & \textbf{0.066} \\*
 & & $\mathcal{L}_{\mathrm{LM}}$ & 0.034 & 0.017 & \textbf{0.012} & \textbf{0.012} & 0.018 & 0.034 & \textbf{0.017} & 0.077 & \textbf{0.018} \\
Irish & L & $D_{\mathrm{KL}}$ & 0.316 & \textbf{0.281} & 0.282 & 0.299 & 0.319 & 0.310 & \textbf{0.278} & \textbf{0.313} & 0.316 \\*
 & & $\mathcal{L}_{\mathrm{LM}}$ & 0.162 & \textbf{0.158} & 0.171 & 0.177 & 0.182 & 0.154 & \textbf{0.148} & \textbf{0.152} & 0.159 \\
\midrule
Mandarin & H & $D_{\mathrm{KL}}$ & 0.252 & 0.188 & \textbf{0.187} & 0.191 & 0.197 & 0.252 & \textbf{0.188} & 0.146 & \textbf{0.140} \\*
 & & $\mathcal{L}_{\mathrm{LM}}$ & 0.159 & \textbf{0.105} & 0.112 & 0.113 & 0.113 & 0.163 & \textbf{0.106} & 0.094 & \textbf{0.081} \\
Cantonese & M & $D_{\mathrm{KL}}$ & 0.224 & 0.170 & \textbf{0.167} & 0.171 & 0.179 & 0.223 & \textbf{0.171} & 0.144 & \textbf{0.140} \\*
 & & $\mathcal{L}_{\mathrm{LM}}$ & 0.124 & \textbf{0.084} & 0.089 & 0.090 & 0.092 & 0.123 & \textbf{0.084} & 0.032 & \textbf{0.031} \\
Burmese & M & $D_{\mathrm{KL}}$ & 0.167 & \textbf{0.056} & 0.058 & 0.058 & 0.058 & 0.168 & \textbf{0.057} & 0.085 & \textbf{0.053} \\*
 & & $\mathcal{L}_{\mathrm{LM}}$ & 0.150 & \textbf{0.036} & 0.039 & 0.037 & \textbf{0.036} & 0.148 & \textbf{0.036} & 0.034 & \textbf{0.004} \\
Tibetan & L & $D_{\mathrm{KL}}$ & 0.056 & \textbf{0.032} & 0.035 & 0.035 & 0.037 & 0.058 & \textbf{0.033} & 0.040 & \textbf{0.027} \\*
 & & $\mathcal{L}_{\mathrm{LM}}$ & 0.009 & \textbf{0.000} & 0.002 & \textbf{0.000} & 0.003 & 0.011 & \textbf{0.001} & 0.005 & \textbf{$-$0.005} \\
\midrule
Arabic & H & $D_{\mathrm{KL}}$ & 0.170 & \textbf{0.130} & 0.134 & 0.136 & 0.136 & 0.167 & \textbf{0.131} & 0.094 & \textbf{0.088} \\*
 & & $\mathcal{L}_{\mathrm{LM}}$ & 0.127 & \textbf{0.101} & 0.104 & 0.106 & 0.105 & 0.124 & \textbf{0.102} & 0.042 & \textbf{0.032} \\
Hebrew & M & $D_{\mathrm{KL}}$ & 0.204 & \textbf{0.156} & \textbf{0.156} & 0.162 & 0.165 & 0.205 & \textbf{0.157} & 0.100 & \textbf{0.094} \\*
 & & $\mathcal{L}_{\mathrm{LM}}$ & 0.081 & \textbf{0.055} & 0.060 & 0.060 & 0.062 & 0.083 & \textbf{0.056} & 0.041 & \textbf{0.035} \\
Amharic & M & $D_{\mathrm{KL}}$ & 0.075 & 0.044 & \textbf{0.042} & 0.044 & 0.049 & 0.075 & \textbf{0.044} & 0.045 & \textbf{0.032} \\*
 & & $\mathcal{L}_{\mathrm{LM}}$ & \textbf{$-$0.013} & $-$0.010 & $-$0.008 & $-$0.009 & $-$0.010 & \textbf{$-$0.014} & $-$0.011 & $-$0.008 & \textbf{$-$0.013} \\
Hausa & M & $D_{\mathrm{KL}}$ & 0.119 & 0.115 & \textbf{0.096} & 0.110 & 0.125 & 0.120 & \textbf{0.116} & \textbf{0.131} & \textbf{0.131} \\*
 & & $\mathcal{L}_{\mathrm{LM}}$ & $-$0.092 & $-$0.086 & $-$0.050 & $-$0.064 & \textbf{$-$0.095} & \textbf{$-$0.094} & $-$0.087 & $-$0.077 & \textbf{$-$0.081} \\
\midrule
Indonesian & H & $D_{\mathrm{KL}}$ & 0.113 & \textbf{0.097} & 0.103 & 0.102 & 0.102 & 0.113 & \textbf{0.098} & 0.126 & \textbf{0.123} \\*
 & & $\mathcal{L}_{\mathrm{LM}}$ & 0.073 & \textbf{0.062} & 0.066 & 0.067 & 0.066 & 0.072 & \textbf{0.062} & 0.013 & \textbf{0.009} \\
Malay & M & $D_{\mathrm{KL}}$ & 0.149 & 0.130 & \textbf{0.129} & 0.131 & 0.138 & 0.148 & \textbf{0.130} & \textbf{0.158} & \textbf{0.158} \\*
 & & $\mathcal{L}_{\mathrm{LM}}$ & 0.108 & 0.094 & \textbf{0.092} & 0.095 & 0.102 & 0.106 & \textbf{0.096} & 0.039 & \textbf{0.038} \\
Filipino & M & $D_{\mathrm{KL}}$ & 0.136 & 0.119 & \textbf{0.118} & 0.123 & 0.131 & 0.136 & \textbf{0.120} & 0.145 & \textbf{0.140} \\*
 & & $\mathcal{L}_{\mathrm{LM}}$ & 0.051 & \textbf{0.050} & 0.057 & 0.055 & 0.058 & \textbf{0.051} & \textbf{0.051} & 0.000 & \textbf{$-$0.008} \\
\midrule
Tamil & H & $D_{\mathrm{KL}}$ & 0.115 & 0.042 & \textbf{0.040} & 0.042 & 0.045 & 0.087 & \textbf{0.042} & 0.137 & \textbf{0.045} \\*
 & & $\mathcal{L}_{\mathrm{LM}}$ & 0.078 & 0.019 & \textbf{0.018} & 0.021 & 0.019 & 0.050 & \textbf{0.018} & 0.099 & \textbf{0.005} \\
Telugu & M & $D_{\mathrm{KL}}$ & 0.168 & 0.050 & \textbf{0.045} & 0.047 & 0.052 & 0.135 & \textbf{0.050} & 0.197 & \textbf{0.053} \\*
 & & $\mathcal{L}_{\mathrm{LM}}$ & 0.130 & 0.027 & \textbf{0.020} & 0.022 & 0.027 & 0.095 & \textbf{0.026} & 0.162 & \textbf{0.024} \\
Malayalam & M & $D_{\mathrm{KL}}$ & 0.133 & 0.042 & \textbf{0.037} & 0.040 & 0.045 & 0.116 & \textbf{0.042} & 0.183 & \textbf{0.050} \\*
 & & $\mathcal{L}_{\mathrm{LM}}$ & 0.096 & 0.017 & \textbf{0.015} & 0.017 & 0.018 & 0.079 & \textbf{0.016} & 0.131 & \textbf{0.006} \\
\midrule
Turkish & H & $D_{\mathrm{KL}}$ & 0.147 & \textbf{0.128} & 0.136 & 0.136 & 0.134 & 0.148 & \textbf{0.129} & 0.134 & \textbf{0.133} \\*
 & & $\mathcal{L}_{\mathrm{LM}}$ & 0.069 & \textbf{0.057} & 0.070 & 0.071 & 0.065 & 0.070 & \textbf{0.059} & \textbf{0.018} & \textbf{0.018} \\
Azerbaijani & M & $D_{\mathrm{KL}}$ & 0.127 & 0.112 & \textbf{0.106} & 0.111 & 0.118 & 0.127 & \textbf{0.112} & 0.123 & \textbf{0.122} \\*
 & & $\mathcal{L}_{\mathrm{LM}}$ & 0.028 & \textbf{0.022} & 0.026 & 0.026 & 0.031 & 0.029 & \textbf{0.022} & $-$0.015 & \textbf{$-$0.019} \\
Kazakh & M & $D_{\mathrm{KL}}$ & 0.096 & 0.081 & \textbf{0.080} & 0.081 & 0.085 & 0.095 & \textbf{0.081} & 0.080 & \textbf{0.078} \\*
 & & $\mathcal{L}_{\mathrm{LM}}$ & 0.028 & \textbf{0.020} & 0.026 & 0.021 & 0.023 & 0.026 & \textbf{0.019} & $-$0.024 & \textbf{$-$0.025} \\
Uyghur & L & $D_{\mathrm{KL}}$ & 0.119 & 0.102 & \textbf{0.089} & 0.096 & 0.110 & 0.119 & \textbf{0.102} & 0.089 & \textbf{0.084} \\*
 & & $\mathcal{L}_{\mathrm{LM}}$ & \textbf{$-$0.013} & $-$0.012 & $-$0.003 & $-$0.005 & $-$0.008 & \textbf{$-$0.014} & $-$0.012 & $-$0.027 & \textbf{$-$0.029} \\
\midrule
Vietnamese & H & $D_{\mathrm{KL}}$ & 0.163 & 0.131 & \textbf{0.130} & 0.134 & 0.136 & 0.163 & \textbf{0.131} & \textbf{0.102} & \textbf{0.102} \\*
 & & $\mathcal{L}_{\mathrm{LM}}$ & 0.121 & \textbf{0.097} & \textbf{0.097} & 0.105 & 0.103 & 0.122 & \textbf{0.096} & 0.034 & \textbf{0.032} \\
Khmer & M & $D_{\mathrm{KL}}$ & 0.112 & \textbf{0.047} & 0.048 & 0.050 & 0.051 & 0.112 & \textbf{0.048} & 0.116 & \textbf{0.053} \\*
 & & $\mathcal{L}_{\mathrm{LM}}$ & 0.045 & \textbf{0.005} & 0.010 & 0.009 & 0.008 & 0.046 & \textbf{0.005} & 0.029 & \textbf{$-$0.021} \\*
\midrule
Mean & & $R$ & 1.000 & 0.677 & \textbf{0.665} & 0.689 & 0.722 & 1.000 & \textbf{0.697} & 1.000 & \textbf{0.767} \\*
 & & $\bar\delta$ & 0.071 & \textbf{0.045} & 0.050 & 0.049 & 0.049 & 0.068 & \textbf{0.045} & 0.042 & \textbf{0.020} \\
\bottomrule
\end{longtable}
\endgroup

\section{SoftCKA details} \label{app:softcka}

\paragraph{Token representations and kernels.}
For each parallel sentence pair $s \in \mathcal{D}$, let
$\mX_a^{(s)} \in \mathbb{R}^{T_a^{(s)} \times d}$ contain the
token-level hidden states for language $a \in \{1,2\}$ at the
model location under analysis. Let $\mX_{a,i}^{(s)}$ denote the
hidden state of token $i$. We construct the RBF kernel matrices
\begin{equation}
    \mK_{ab}^{(s)}[i,j]
    =
    \exp\!\left(
        -\frac{
            \lVert \mX_{a,i}^{(s)}-\mX_{b,j}^{(s)} \rVert_2^2
        }{2\sigma^2}
    \right),
    \label{eq:softcka-rbf}
\end{equation}
where $a,b \in \{1,2\}$, $1 \leq i \leq T_a^{(s)}$, and
$1 \leq j \leq T_b^{(s)}$. The bandwidth $\sigma > 0$ is shared
across the three matrices $\mK_{11}^{(s)}$, $\mK_{22}^{(s)}$,
and $\mK_{12}^{(s)}$.

The within-language matrices have dimensions
$T_1^{(s)} \times T_1^{(s)}$ and
$T_2^{(s)} \times T_2^{(s)}$, while the cross-language matrix has
dimensions $T_1^{(s)} \times T_2^{(s)}$. The latter compares
every token in one sentence with every token in its translation,
without requiring equal sequence lengths or token correspondences.
Its entries are similarities, not normalized alignment probabilities.

\paragraph{Centering and squared norms.}
Each kernel is centered separately within its sentence pair:
\begin{equation}
    \widetilde{\mK}_{ab}^{(s)}
    =
    \mH_a^{(s)} \mK_{ab}^{(s)} \mH_b^{(s)},
    \qquad
    \mH_a^{(s)}
    =
    \mI_{T_a^{(s)}}
    -
    \frac{1}{T_a^{(s)}}\bm{1}\bm{1}^{\top}.
    \label{eq:softcka-centering}
\end{equation}
This subtracts each entry's row and column means and adds back
the overall matrix mean. We then compute
\begin{equation}
    h_{ab}^{(s)}
    =
    \lVert \widetilde{\mK}_{ab}^{(s)} \rVert_F^2
    =
    \sum_{i=1}^{T_a^{(s)}}
    \sum_{j=1}^{T_b^{(s)}}
    \left(\widetilde{\mK}_{ab}^{(s)}[i,j]\right)^2.
    \label{eq:softcka-energy}
\end{equation}
Thus, $h_{ab}^{(s)}$ measures variation remaining after row and
column mean similarities are removed, rather than the overall
level of token similarity.

\paragraph{Corpus aggregation.}
Equation~\ref{eq:softcka} sums the quantities $h_{ab}^{(s)}$
over sentence pairs before normalization. It therefore does
not average sentence-level SoftCKA scores. No explicit
length normalization is applied: the cross-language term
contains $T_1^{(s)}T_2^{(s)}$ squared entries, while the
within-language terms contain $(T_1^{(s)})^2$ and
$(T_2^{(s)})^2$ entries.

Consequently, each sentence pair's contribution depends on
both its sequence lengths and its centered kernel values.
When the two token counts are comparable and the mean squared
centered entries remain similar, contributions to these sums
grow approximately quadratically with sequence length.
The normalized corpus score is defined when its denominator
is nonzero.

\paragraph{Relationship to CKA.}
Standard empirical HSIC is computed from two Gram matrices
$\mK,\mL \in \mathbb{R}^{N \times N}$ indexed by the same
$N$ paired observations:
\begin{equation}
    \operatorname{HSIC}(\mK,\mL)
    =
    \frac{1}{(N-1)^2}
    \operatorname{tr}(\mK\mH\mL\mH),
    \qquad
    \mH = \mI_N - \frac{1}{N}\bm{1}\bm{1}^{\top}.
\end{equation}
CKA normalizes this quantity by the corresponding self-HSIC
terms \citep{pmlr-v97-kornblith19a}. SoftCKA adopts a similar
normalization structure, but its cross-language term is
$\lVert \mH_1^{(s)}\mK_{12}^{(s)}\mH_2^{(s)} \rVert_F^2$.
This term is not the standard HSIC estimator over paired tokens.

Because $\mK_{12}^{(s)}$ uses distances between hidden states
from different languages, SoftCKA compares them in their shared
representation space. Its interpretation is normalized centered
kernel similarity; it does not recover or verify semantic
token correspondences.
\section{SoftCKA of Block Deltas} \label{softcka_delta}

Our primary metric looks at the hidden state entering each layer, \verb+attn_in+. However, we find that a slight modification reveals interesting results. As discussed in Section~\ref{sec:observation}, we also calculate SoftCKA of the residual-stream contribution of each block. Since this treats the block as a black box, it is comparable across various attention blocks and even the MLP block. In Figure~\ref{diagram}, these values are \verb+attn_delta+ and \verb+mlp_delta+.

The following visualization for Qwen3.5 shows the sequential progress through both the attention and MLP blocks. On top of the major ``event'' happening at the first full attention layer, it also displays the ``preparation'' that occurs in the few preceding blocks.

\begin{figure*}[h]
    \centering
    \includegraphics[width=\linewidth]{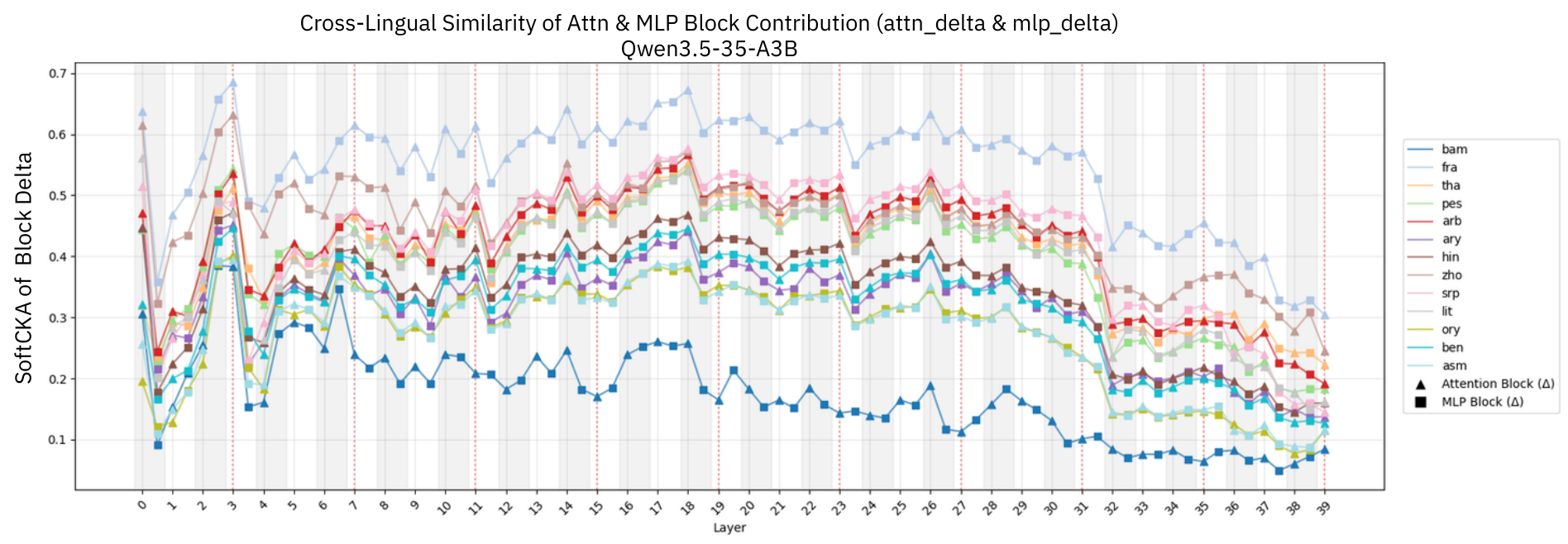}
    \caption{SoftCKA of the residual-stream contributions from the attention and MLP blocks in Qwen3.5. The blocks preceding the first full-attention layer show a progressive change before the pronounced event at that layer. Similarity has a massive drop right after this layer.}
    \label{fig:qwen35-all-delta}
\end{figure*}
\section{Comparison to SWA Hybrids} \label{swa_hybrid}

We elaborate here upon Finding 5. The visualization for \olmo-3, a SWA-hybrid, is in Figure~\ref{fig:vis-jagged}, while for Tiny Aya Global, it is here below.

\begin{figure}[h]
    \centering
    \includegraphics[width=0.7\linewidth]{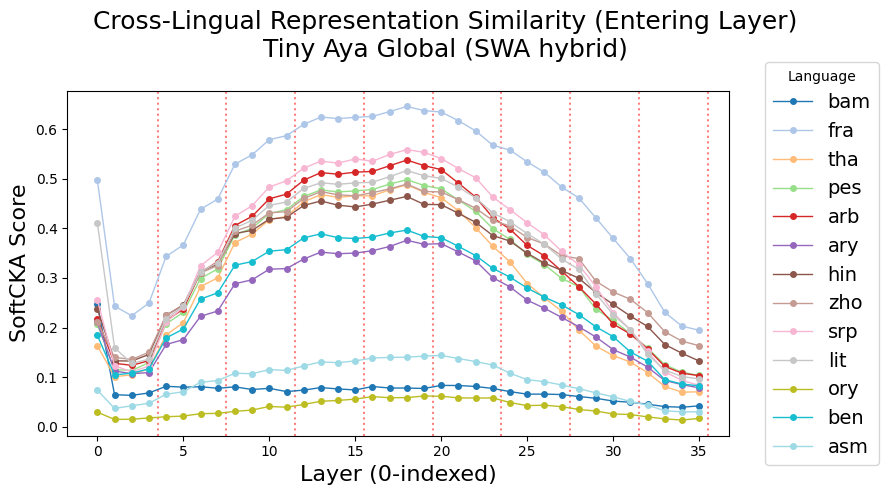}
    \caption{SoftCKA for Tiny Aya Global, which is an SWA-hybrid model. Red dotted lines mark full-attention layers.}
    \label{tinyaya_layers}
\end{figure}

Interestingly, it is found that positional embeddings are not useful in the SWA layers \citep{yang2025rope,qiao2026rethinkingroleefficientattention}, so numerous new such hybrid models do not use it outside of full attention \citep{tinyaya,kimi3}. We find no evidence that this significantly impacts cross-lingual representations.
\section{MoE Routing Alignment}
\label{moe_vis}

For the mixture-of-experts Qwen and Ring model pairs, we compute the routing-divergence metric of \citet{bandarkar2026multilingual}. The metric measures cross-lingual MoE routing agreement at the sequence level using parallel sentences. For each sequence, the router weights are mean-pooled across tokens and then we take the Jensen-Shannon divergence between the resulting distributions for the two languages. We apply this metric to parallel samples from FloRES \citep{nllb2022} and report corpus-wide averages at each layer. Lower JS divergence indicates more similar expert-routing behavior across languages.

\begin{figure*}[h]
  \centering
  \includegraphics[width=0.68\textwidth]{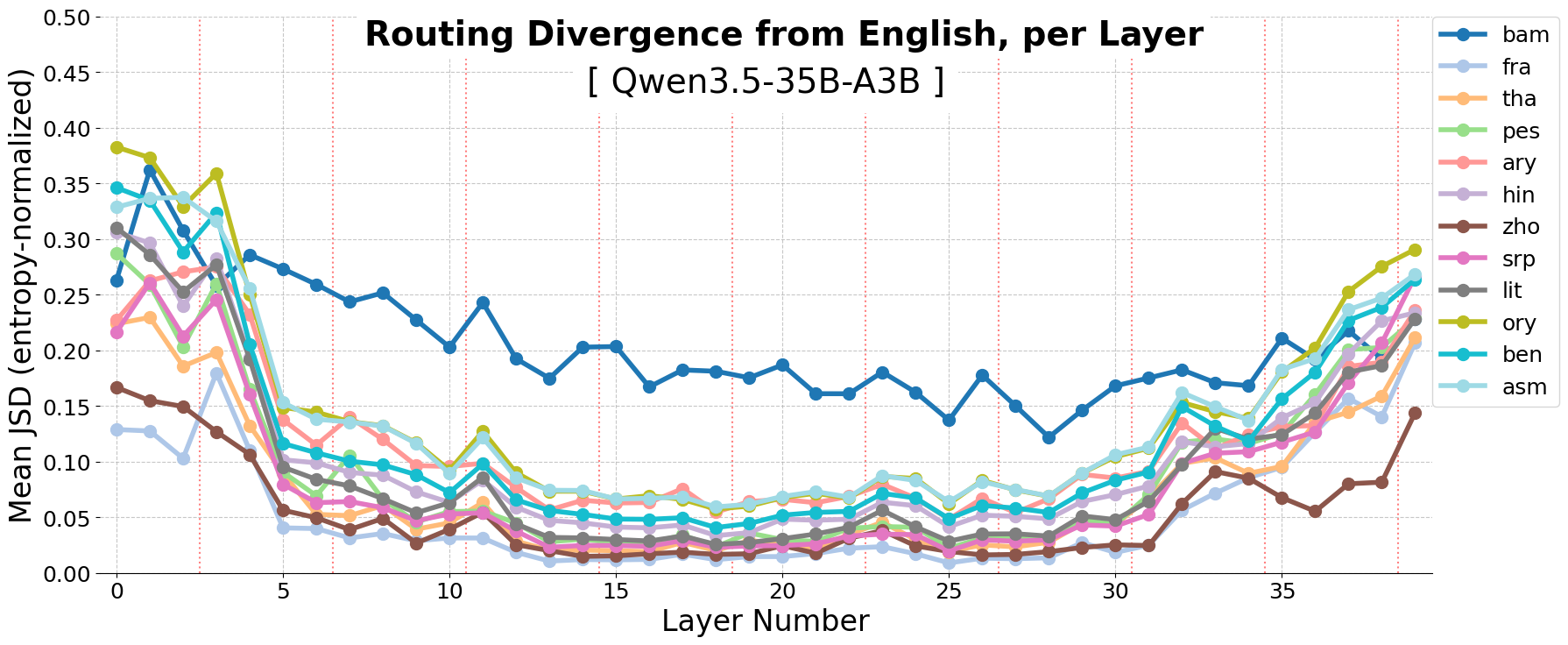}
  \caption{Layer-wise cross-lingual MoE routing divergence for Qwen3.5 (hybrid). Lower values indicate more similar routing across languages. Red dotted lines mark the full attention layers.}
  \label{fig:qwen35-moe-router-jsd}
\end{figure*}

\begin{figure*}[h]
  \centering
  \includegraphics[width=0.68\textwidth]{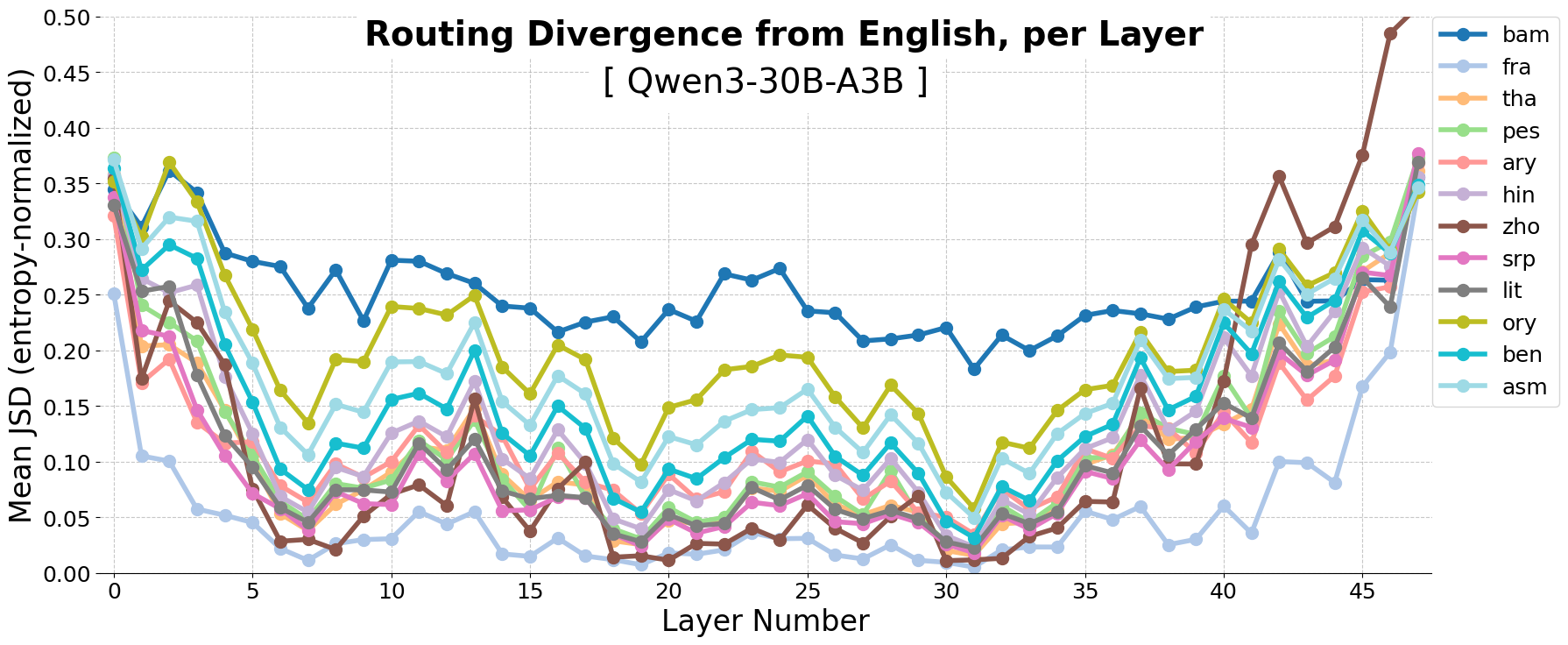}
  \caption{Layer-wise cross-lingual MoE routing divergence for Qwen3 (non-hybrid).}
  \label{fig:qwen3-moe-router-jsd}
\end{figure*}

\begin{figure*}[h]
  \centering
  \includegraphics[width=0.68\textwidth]{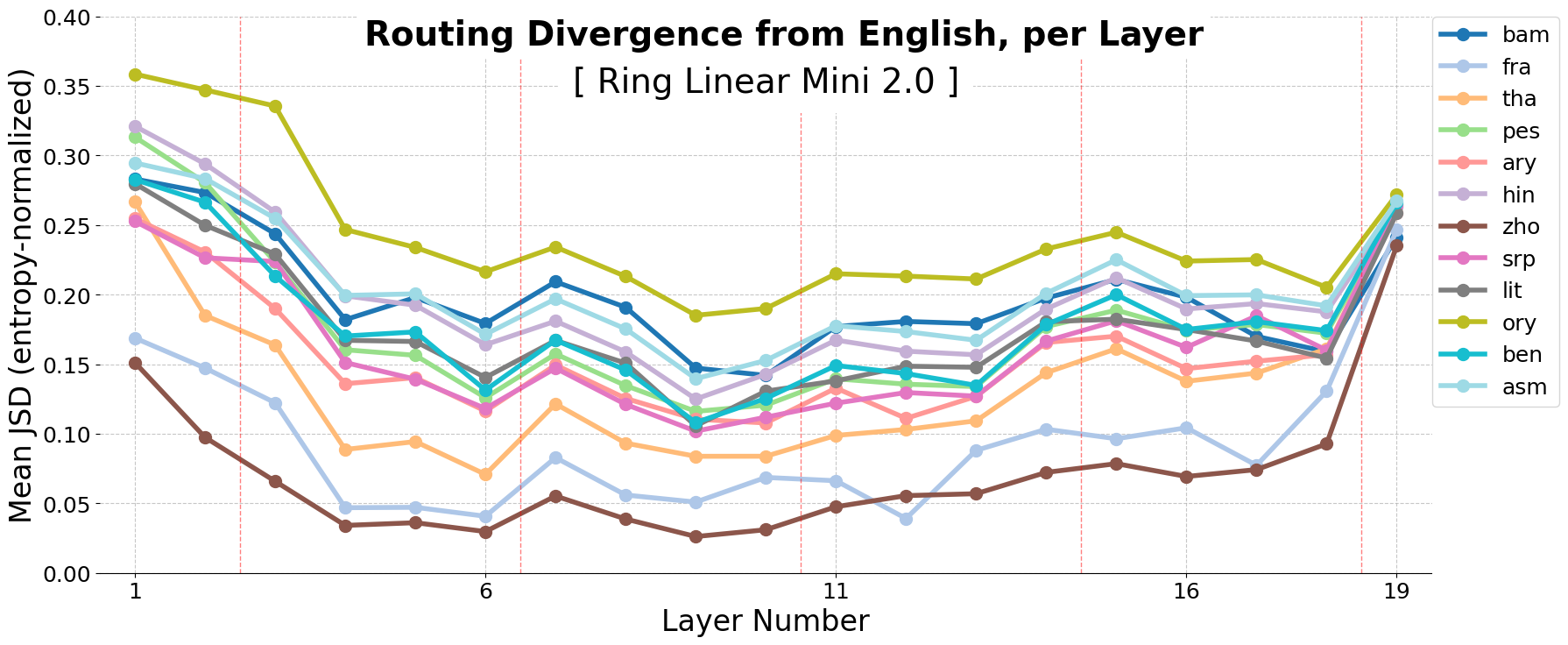}
  \caption{Layer-wise cross-lingual MoE routing divergence for Ring-mini-linear-2.0 (hybrid). Red dotted lines mark the full attention layers.}
  \label{fig:ringlinear-moe-router-jsd}
\end{figure*}

\begin{figure*}[h]
  \centering
  \includegraphics[width=0.68\textwidth]{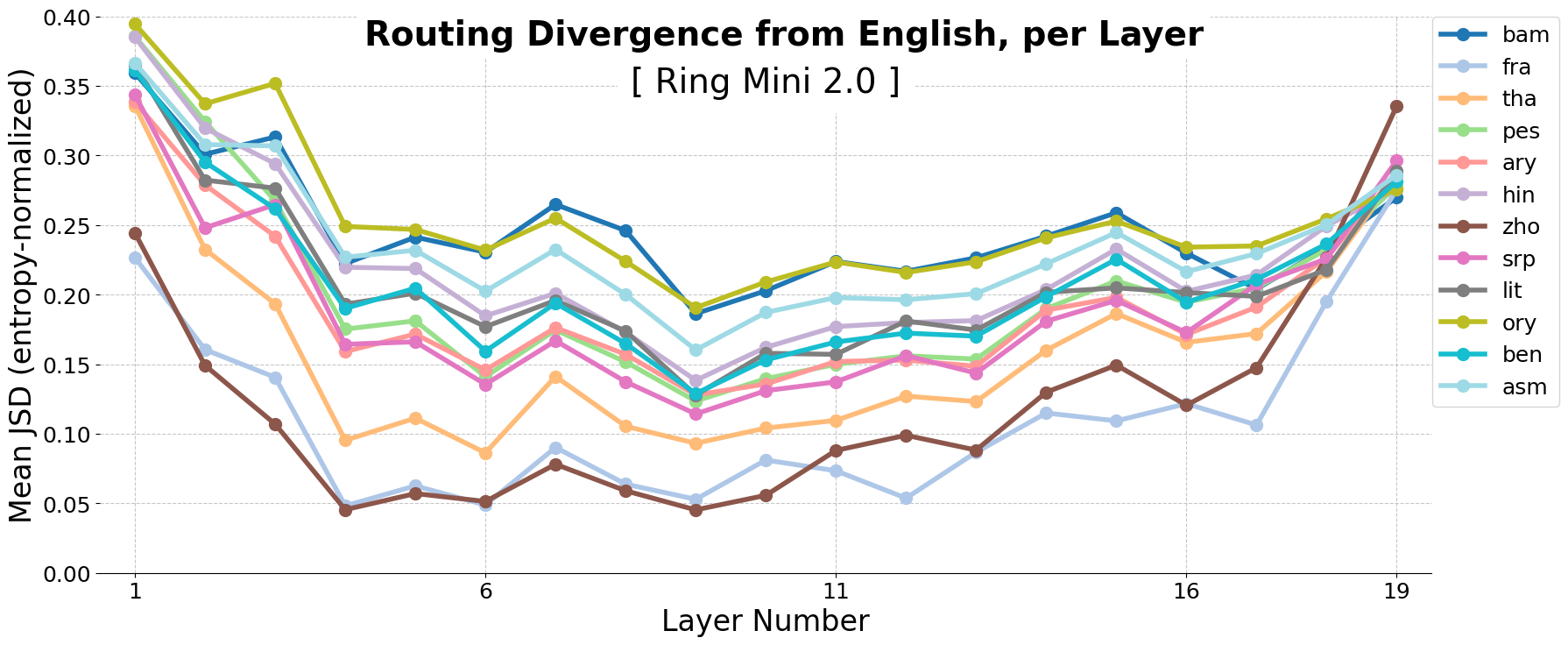}
  \caption{Layer-wise cross-lingual MoE routing divergence for Ring-mini-2.0 (non-hybrid).}
  \label{fig:ring-moe-router-jsd}
\end{figure*}

Figures~\ref{fig:qwen35-moe-router-jsd}-\ref{fig:ring-moe-router-jsd} show the layer-wise results. In the hybrid models, Qwen3.5-35B-A3B and Ring-mini-linear-2.0, routing divergence remains relatively flat through the initial recurrent layers and begins to decrease only after the first full-attention layer. By contrast, divergence begins decreasing immediately in the non-hybrid Qwen3-30B-A3B and Ring-mini-2.0 models.

\end{document}